\documentclass[10pt,leqno]{article}
\usepackage{bbding}
\usepackage{pifont}
\usepackage{multicol, multirow}
\usepackage{subfigure}
\usepackage{threeparttable}
\usepackage{graphicx}
\usepackage[T1]{fontenc}
\usepackage{authblk}
\usepackage{indentfirst,csquotes}
\usepackage{blindtext}
\usepackage{amssymb,amsthm,amsmath}
\usepackage{xcolor,paralist,hyperref}
\usepackage{fancyhdr,etoolbox}
\usepackage{booktabs}
\usepackage{microtype}
\usepackage{utfsym}

\newcommand\keywords[1]{\textbf{Keywords}: #1}

\hypersetup{ colorlinks=true, linkcolor=black, filecolor=black, urlcolor=black }

\title{Transformer-based Tooth Alignment Prediction with Occlusion and Collision Constraints} 
\author[1]{ZhenXing Dong}
\author[1]{JiaZhou Chen}
\author[2]{YangHui Xu}
\affil[1]{\textit{College of Computer Science and Technology College of Software, ZheJiang University of Technology}}
\affil[ ]{201806062003@zjut.edu.cn, cjz@zjut.edu.cn, 1111912014@zjut.edu.cn}
\date{} 

\begin{document}

\maketitle
\let\thefootnote\relax

\begin{abstract}
    The planning of digital orthodontic treatment requires providing tooth alignment, which not only consumes a lot of time and labor to determine manually but also relays clinical experiences heavily. In this work, we proposed a lightweight tooth alignment neural network based on Swin-transformer. We first re-organized 3D point clouds based on virtual arch lines and converted them into order-sorted multi-channel textures, which improves the accuracy and efficiency simultaneously. We then designed two new occlusal loss functions that quantitatively evaluate the occlusal relationship between the upper and lower jaws. They are important clinical constraints, first introduced to the best of our knowledge, and lead to cutting-edge prediction accuracy. To train our network, we collected a large digital orthodontic dataset that has 855 clinical cases, including various complex clinical cases. This dataset will benefit the community after its release since there is no open dataset so far. Furthermore, we also proposed two new orthodontic dataset augmentation methods considering tooth spatial distribution and occlusion. We evaluated our method with this dataset and extensive experiments, including comparisons with STAT methods and ablation studies, and demonstrate the high prediction accuracy of our method.
\end{abstract} 


\bigskip

\noindent

\section{Introduction}
Tooth correction, medically known as orthodontics, primarily involves the use of metal braces \cite{eliades2008manufacturing} or clear aligners \cite{kumar2018invisalign} to alleviate or rectify the conditions of dental misalignment and malformation. However, regardless of the type of appliance employed, the alignment of teeth constitutes a crucial preliminary step in orthodontic treatment planning. With the widespread adoption of digital acquisition technologies, computer-aided alignment design has been paid extensive attentions, such as those based on intraoral scanners \cite{revilla2023overview} and cone-beam computed tomography (CBCT) \cite{cheng2015personalized}. Three-dimensional tooth models are initially segmented individually, and then repositioned by the clinician considering various alignment factors such as the extent of dental protrusion, dental skeletal relationship, and periodontal conditions of the patient and etc. It heavily relies on the clinical expertise of orthodontists and is time-consuming, thereby significantly increases the duration and cost of orthodontic treatment planning.

With the advancement of artificial intelligence, learning-based methods for tooth alignment are emerging rapidly \cite{lee2024applications}, aiming at achieving fully-automated tooth alignment. Among these methods, PointNet-based ones are particularly representative \cite{tanet2020}\cite{PSTN2020}\cite{iorthopredictor2020}\cite{lei2023automatic}. TANet \cite{tanet2020} employs PointNet to construct a feature extraction module, encoding both global information of the jaw and local information of teeth, and utilizes MLP to design regressors for predicting the position of each tooth. PSTN \cite{PSTN2020} utilizes both PointNet \cite{qi2017pointnet} and PointNet++ \cite{qi2017pointnet++} for feature encoding, refining features based on a combination of local and global latent vectors to regress tooth transformation parameters. TAligNet \cite{iorthopredictor2020}, also based on PointNet encoders and MLP decoders, employs Squeeze-and-Excitation Blocks and shared FC sequences for feature propagation to predict alignment parameters.

PointNet-based tooth alignment prediction mehtods showed great potentials, but limitations are revealed in representing local features of point clouds \cite{wang2019dynamic} recently. Point Transformer introduces transformer architecture into the feature extraction of three-dimensional point clouds. Compared to PointNet, it offers a larger receptive field, thus enabling better feature extraction. Point Transformer has evolved to its third version, which employs various serialization techniques on three-dimensional point clouds, further expanding the receptive field to enhance accuracy and efficiency. Drawing initial inspiration, this paper introduces a more advanced shift window transformer (referred to as Swin-T). It incorporates sliding window operations and hierarchical merging design on the foundation of traditional vision transformers, addressing issues such as lower precision due to the variability of objects, excessive pixel count leading to high computational complexity, and low computational efficiency encountered in transformer-based methods. As teeth share similar sizes and structures, they are sampled into uniformly sized three-dimensional point clouds and transformed into regular multi-channel data, forming ordered data. Building upon this data organization, this paper proposes a multi-level channel compression structure based on Swin-T (SWTBS) and an SWTP module to respectively extract global information of tooth centers and local information of tooth point clouds. Benefiting from the performance optimization of shift windows and communication between windows, features of individual teeth can mutually inform one another, gradually expanding the receptive field and exhibiting excellent global control over entire dental arches, thereby achieving higher prediction accuracy.

Existing learning-based tooth alignment methods primarily consider reconstruction error and transformation parameter error when designing constraint terms. For instance, PSTN and TAligNet focus on the spatial discrepancies between predicted results and ground truth of individual tooth point clouds, as well as the numerical deviations of transformation matrices. \cite{wang2022tooth} further incorporates landmark constraints, simulating Andrew's six orthodontic factors through key points. Among these, the occlusal relationship between the upper and lower jaws is the most crucial factor. \cite{tanet2020} also includes nodes of upper and lower teeth in the calculation set for chamfer vector loss, while \cite{lei2023automatic} considers the relative positions of the center points of upper and lower teeth in the relative position loss. These can be regarded as simple evaluations of the occlusal relationship between upper and lower jaws. However, occlusion involves complex phenomena from both biological and physical disciplines. Methods based on angles or key points suffer from significant errors and have limited effectiveness. Therefore, this paper introduces two additional precise occlusal loss functions for upper and lower teeth occlusion, building upon the existing model reconstruction loss and rotation-translation transformation parameter loss:

\begin{itemize}
    \item Occlusal Projecting Overlap: The occlusal projecting overlap of a tooth is defined as follows: from a top-down perspective, it encompasses the overlap between the crown region of the tooth in question and the crown region of the corresponding tooth in the opposing jaw. We quantify the disparity between the predicted and ground truth occlusal projecting overlaps to evaluate the alignment of occlusion. Detailed explanations of occlusal projecting overlap and the corresponding calculation of the loss function are provided in Section \ref{subsubsection:Occlusal projecting overlap}.

    \item Occlusal Distance Uniformity: Ideally, there should be good concave-convex contact relationships between the buccal cusps, mesiobuccal cusps, and mesiopalatal concavities of the upper and lower molars \cite{andrews1972six}. For posterior teeth, the occlusal uniformity can be expressed as the uniformity of vertical distances between points within the occlusal projecting overlap of tooth and the corresponding area of the opposing jaw. For anterior teeth, occlusal uniformity primarily focuses on the relative positions and covering relationships between upper and lower anterior teeth. Normal anterior tooth occlusion entails proper overlapping of upper anterior teeth over lower anterior teeth, with corresponding and aligned positions between upper and lower anterior teeth.
\end{itemize}

Regarding training data, existing methods suffer from both small volume and lack of public availability. This paper has gathered a dataset of 855 orthodontic alignment plans, which will be publicly released along with the publication of this paper, alongside the source code. To further augment the dataset, we have devised a method of data augmentation: constraint augmentation during the preprocessing stage. The effectiveness of this data augmentation techniques has also been evaluated in our experiments. Additionally, we directly collected data where dentists manually aligned the segmented initial tooth models. Compared to intraoral scan data before and after orthodontic used in previous methods, our data offers two significant advantages: (1) The aligned label data correspond one-to-one with the original intraoral scan tooth model data in terms of point cloud positions, eliminating the need for additional point cloud matching computations to align posture changes before and after orthodontic. Given that parts of the tooth crown extend beyond the gum line may change, this matching process is not straightforward or stable; (2) Based on interviews with multiple orthodontists, it is known that patients' orthodontic treatment outcomes do not always fully meet their expectations. Therefore, for orthodontists, using the target position manually aligned by dentists as the ground truth for tooth alignment is often more reasonable than post-orthodontic intraoral scan data.

To summarize, our contributions are as follows:

\begin{itemize}
    \item A lightweight tooth alignment network based on Swin-T is designed to replace traditional three-dimensional point cloud feature extraction encoders. It organizes scattered point clouds into regularly sized and orderly sorted multi-channel texture forms, ensuring high efficiency and seamless compatibility with complex scenarios such as missing teeth and wisdom teeth, surpassing the accuracy of the STAT method in tooth alignment.
    \item Two occlusal loss functions, namely the occlusal projecting overlap loss and occlusal distance uniformity loss (comprising anterior and posterior tooth uniformity), are designed based on medical domain knowledge. These functions enable more accurate and efficient quantitative evaluation of the occlusal relationship between the upper and lower jaws.
    \item An extensively annotated orthodontic alignment dataset, tailored to better suit the requirements of orthodontists, has been labeled. Plans are underway to make this dataset publicly available to address the issue of scarcity of public data in the field. Additionally, two point cloud data augmentation methods specific to tooth alignment are proposed, and the impact of data augmentation proportions on final prediction results is explored through ablation experiments.
\end{itemize}

\section{Related Works}

\subsection{learning-based tooth alignment}
Existing AI tooth alignment methods primarily use three-dimensional point cloud data as input, rather than mesh or voxel data. Early AI tooth alignment methods were mainly based on the PointNet\cite{qi2017pointnet} structure and its derivatives. PointNet\cite{qi2017pointnet}, proposed by Charles R. Qi et al. from Stanford University, is a neural network model capable of directly processing three-dimensional point clouds. It learns the spatial encoding of each point in the input point cloud and then utilizes the features of all points to obtain a global point cloud feature. Although PointNet is proficient at extracting global features, its ability to extract local features is limited, and its capability to handle details and generalize to complex scenes is quite restricted. Therefore, PointNet++ \cite{qi2017pointnet++} introduced a hierarchical feature extraction structure to effectively extract both local and global features. Based on the point cloud feature extraction capabilities of \cite{qi2017pointnet} or \cite{qi2017pointnet++}, TANet \cite{tanet2020} utilizes PointNet \cite{qi2017pointnet} to encode the point cloud features of intraoral scan segmentation models, including global and local features. It then employs graph neural networks to connect and communicate tooth local features, regressing the 6dof information of teeth. PSTN \cite{PSTN2020} uses PointNet \cite{qi2017pointnet} to encode global and local features and PointNet++ \cite{qi2017pointnet++} to encode local features. After fusion, it uses a decoder designed based on PointNet \cite{qi2017pointnet} to regress orthodontic transformations of teeth. TAligNet \cite{iorthopredictor2020} achieves feature extraction of three-dimensional tooth models and tooth arrangement. It utilizes PointNet \cite{qi2017pointnet} as the feature encoder and employs fully connected layer sequences and SE blocks for feature propagation, finally using fully connected layers to regress rotation and translation.

In addition to using PointNet, recent AI tooth alignment methods have also adopted emerging network structures such as DGCNN and diffusion models. Wang et al.\cite{wang2022tooth} proposed an improvement to TANet using tooth landmarks, where DGCNN \cite{wang2019dynamic} was utilized to extract point cloud information. They proposed a hierarchical regression using a three-layer graph neural network structure to better predict the displacement transformation of each tooth. Additionally, they innovatively encoded dental representation information, such as the tooth latent shape code and tooth frame, based on initial and target states of the teeth, then applied conditional diffusion to predict the orthodontic path distribution from the initial to the target state, achieving remarkable results\cite{fan2024collaborative}, where the landmark serves as a component of the tooth frame. Lei et al.\cite{lei2023automatic} also employed diffusion models; they used probabilistic diffusion models to iteratively denoise random variables, learning the distribution of transformation matrices for dental transitions from malocclusion to normal occlusion, thus achieving more realistic orthodontic predictions. It should be noted that this method, to fully utilize effective features, employs different encoding networks to extract both local tooth and global gingival features, simultaneously utilizing mesh and point cloud representations. Furthermore, the network structure of TAligNet mentioned above was actually proposed in image-based AI tooth alignment methods, iOrthoPredictor \cite{iorthopredictor2020}, which utilize three-dimensional geometric information encoded in the unsupervised generative model StyleGAN \cite{karras2019style}. Through meaningful paths in latent space normals, alignment processes in image space are generated.

In terms of loss function design, reconstruction error is the primary consideration in existing methods. TANet \cite{tanet2020}, TAligNet \cite{iorthopredictor2020}, landmark \cite{wang2022tooth}, and TADPM \cite{lei2023automatic} all utilize point-wise nearest distance reconstruction error loss functions to assess the spatial distance gap between predicted results and ground truth. TANet \cite{tanet2020} introduced geometric spatial relation loss, defining the point-wise nearest distance between each tooth point cloud and adjacent tooth point clouds (including teeth within the same jaw and relative jaw). The difference between the computed nearest distances on predicted results and ground truth is utilized as the geometric spatial relation loss, which was also applied in landmark \cite{wang2022tooth}. Additionally, starting from the perspective of landmarks, landmark detection loss and landmark target loss were proposed by \cite{wang2022tooth} to evaluate the gap between predicted results and ground truth landmarks. PSTN \cite{PSTN2020} introduced parameterized matrix loss, where the parameterized matrix represents the transformation from the initial position to the aligned position. PSTN \cite{PSTN2020} calculates the difference between predicted parameterized matrices and ground truth as the loss function. TADPM \cite{lei2023automatic} also employs the squared error between predicted transformation matrices and ground truth as the loss for the diffusion model. A similar approach is the transformation parameter loss of landmark \cite{wang2022tooth}, which calculates the rotation and translation parameters from the initial position to the aligned position and compares the predicted parameter values with the ground truth. The open-source code Auto Tooth Arrangement \cite{huang2022github} also utilizes transformation parameter loss functions. The relative position loss of TADPM \cite{lei2023automatic} describes the squared error between the spatial predicted distance and the real distance of the center points of tooth point clouds and adjacent tooth point clouds. However, the occlusal interaction between upper and lower teeth involves both biological and physical disciplines, which are highly complex. Therefore, relying solely on angular or center point calculation methods results in significant errors and limited effectiveness. To address this, this paper designs more accurate occlusal loss functions, including occlusal projecting overlap and occlusal distance uniformity, and implements efficient computational methods.

\subsection{Shift window transformer}
The transformer originated in the field of natural language processing, such as the explosive development seen in recent years with ChatGPT. It has also made significant strides in the field of computer vision, with the Vision Transformer \cite{dosovitskiy2020image} being a classic example. It completely disregards CNN and directly applies the self-attention mechanism to sequences of image patches, performing remarkably well in image classification tasks. Swin-T \cite{liu2021swin}, based on the Vision Transformer \cite{dosovitskiy2020image}, introduces a movable window, constraining the calculation of sub-attention to non-overlapping local windows. Communication between windows ensures effectiveness while bringing about higher efficiency. Swin3D \cite{yang2023swin3d} converts sparse point clouds into voxel data of regular dimensions, enabling the application of the general Swin-T backbone to traditional three-dimensional point cloud tasks. This method utilizes FPS farthest point sampling during point cloud downsampling, pooling neighbor information using KNN for each point. However, this downsampling approach, compared to sliding window-based multi-level feature fusion methods, is more computationally intensive in terms of time and space complexity. Additionally, it completely overlooks the advantage of feature extraction in relative positions between teeth brought about by serialization of three-dimensional point cloud sequences.

The approach in this paper builds upon the multi-level feature fusion network architecture in Swin-T \cite{liu2021swin}. Based on this architecture, we progressively shrink the data size layer by layer using sliding windows as references, reducing spatial and temporal overheads while enhancing global receptive fields. Additionally, we uniformly sample and arrange the data for each tooth to obtain multi-channel data with regular image-like dimensions, thereby avoiding direct utilization of KNN pooling downsampling similar to PTV \cite{zhao2021point}, enhancing information propagation efficiency.

Point Transformer \cite{zhao2021point} introduces the multi-head self-attention mechanism of transformers into three-dimensional point cloud analysis for the first time. It utilizes the k-nearest neighbors of points for self-attention mapping and employs the relative positions between points as natural positional encoding inputs. PTV2 \cite{wu2022point} is designed with grouped weight encoding layers, inheriting the advantages of learnable weight encoding and multi-head attention. It also introduces novel lightweight partition-based pooling methods to achieve better spatial alignment and more efficient sampling. Recent work like PTV3 \cite{wu2023point}, inspired by OCT Transformer \cite{wang2023octformer} and Flat Transformer \cite{liu2023flatformer} which utilize structured point cloud data, arranges unordered point cloud data, introducing the concept of sequences, widening receptive fields and improving performance. However, this generic sorting method does not fully exploit the geometric features implicit in tooth point clouds when extracting tooth features for alignment. To fully consider the specificity of orthodontic tooth alignment tasks, we encode point cloud data through serialization based on dental simulate-arch line, better utilizing the overall morphology of teeth in the oral cavity and achieving a more efficient mechanism for feature exchange and transmission.

\subsection{dataset and data augmentation}
The datasets used in existing neural network orthodontic methods mostly originate from dental clinics and hospitals, and all datasets from these methods are currently unavailable. Table \ref{table1} presents the dataset situations published in major literature that we have compiled. It is evident from the table that, except for TAligNet, the number of datasets for other methods is relatively small, with totals all below 320. Due to the complexity of dental conditions in orthodontic patients, training on these datasets may not be sufficient. The authors of TAligNet have a deep collaboration with dental hospitals, and they possess a large dataset totaling 8995 sets, with a training-to-testing ratio of 8000:995. Due to extensive training, their test results are relatively good. However, it is regrettable that the relevant data has not been made publicly available. In collaboration with orthodontic design teams, we have collected a dataset comprising 855 sets, with 755 sets used for training, 30 sets used for validation, and the rest for testing. We plan to release this dataset after the acceptance of our paper. To the best of our knowledge, it will be the first orthodontic dataset released in the community, which can greatly benefit researchers in the future.

\begin{table}
	\caption{Overview of the dataset for each method}
	\centering
	\begin{tabular}{lllllll}
		\toprule
		Model & Total size & Train & Val & Test & Sample num & Sample func \\
		\midrule
		TANet & 300 & 200 & 30 & 70 & 400 & Randomly sample \\
		PSTN & 206 & 150 & 28 & 28 & 128 & FPS \\
		landmark & 319 & 239 & 30 & 50 & 400 & FPS \\
		TAligNet & 8995 & 8000 & / & 995 & 1024 & Randomly sample \\
		TADPM & 212 & / & / & / & / & / \\
		Our & 855 & 755 & 30 & 70 & 512 & FPS \\
		\bottomrule
	\end{tabular}
	\label{table1}
\end{table}

Since the existing datasets are not publicly available and are small in scale except for TAligNet\cite{iorthopredictor2020}, data augmentation is particularly important for training models. PSTN\cite{PSTN2020} performed an overall rotation on the teeth models before and after orthodontic treatment, but it could not simulate different tooth alignment situations, thus providing limited enhancement effects. Subsequent methods generally enhanced each tooth individually. For example, TAligNet added random noise to the teeth before orthodontic treatment, landmark set a 30\(\%\) probability for each tooth to undergo rotational perturbation within the range of [-60, 60] degrees and translational perturbation within the range of [-1mm, 1mm]. TANet\cite{tanet2020}, on the other hand, performed an overall rotational enhancement and then applied random rotation within the range of [-30, 30] degrees and translation with a Gaussian distribution \(N(0,1)\) to each tooth. However, the teeth before orthodontic treatment are already misaligned, and randomly rotating and moving them could lead to excessive deviation and distortion. Conversely, reducing the rotation angle and translation distance offers limited improvement in model training. Therefore, enhancing the teeth after orthodontic treatment to simulate pre-treatment data can be considered. For instance, TANet applied the same random rotation and translation to post-treatment data as it did to pre-treatment data. TADPM discarded pre-treatment enhancement and only enhanced post-treatment data to simulate and replace pre-treatment data. Experiments with TADPM demonstrated that enhancing post-treatment data to simulate pre-treatment data is very effective. Another advantage of this approach is that it avoids the differences in point cloud composition and geometric structure of the original teeth models before and after orthodontic treatment. However, methods such as TADPM, which enhance teeth individually, face a critical issue: random rotation and translation can result in tooth interpenetration and significant deviation from the tooth axis, affecting the quality of the augmented data. Therefore, this paper introduces constraints such as collision detection and dental arch alignment to address the issues of tooth interpenetration and axis deviation. Ablation experiments not only proved the effectiveness of these constraints in improving the quality of augmented data but also determined the optimal usage ratio of this augmentation method through experiments, achieving the highest prediction accuracy.

\section{Methodology}
\label{section:Methodology}

\subsection{Problem statement}
We segment the patient's intraoral scan model to obtain the gingival point cloud \( G \) and a collection of crown point clouds \( T \) for 32 teeth. There are a maximum of 16 upper teeth and 16 lower teeth, categorized according to the universal tooth naming standard, with the upper teeth numbered from 1 to 16 and the lower teeth numbered from 17 to 32, totaling up to 32 teeth information (cases of missing teeth will be discussed in Section \ref{subsection:Data re-organization}). To ensure training efficiency, we perform farthest point sampling \cite{moenning2003fast} on the tooth point clouds in \( T \), resulting in a point cloud collection for each sampled crown model, denoted as \( P_t \). We sample to the same number of points based on the principle of consistency. Considering the influence of the number of sampled points on network performance and the window merging mechanism for alignment shift windows, we set the sampling quantity parameter \( N \) to 512. \( T \) and \( P_t \) serve as the input data for our method.

\begin{equation}
    T=\{t_i | 1 \leq i \leq 32\}
\end{equation}
\begin{equation}
    P=\{p^{t}_{j} | t \in T, 1 \leq j \leq 512\}
\end{equation}

We define the set of centroid points for each tooth as \( C_t \), where the geometric centroid of the sampled point cloud \( P_t \) corresponding to tooth \( t \) is designated as the centroid \( c_t \) of that tooth.

\begin{equation}
    C_{t}=\{c_{t} | t \in T, c_{t} = \underset{p \in P_{t}}{Ave}(p) \}
\end{equation}

The objective of this study is to predict the 6DoF pose transformation parameters for each tooth model. A detailed overview of the evaluation of 6DoF is provided in \cite{du2021vision}. Typically, among the six parameters of 6DoF, the first three parameters represent the orientation of the model, while the last three parameters denote the position of the model's center in the world coordinate system. Since our dataset lacks local coordinate axis data for teeth, the orientation of teeth in both initial and resulting states is not clearly defined. Therefore, we replace the three parameters representing orientation with axis-angle vectors corresponding to the initial rotation of the tooth model to the ideal alignment position.

To train the network model for prediction, the dataset also includes post-orthodontic ground truth data \( P^*_t \), where the points correspond one-to-one with those in the pre-orthodontic data \( P_t \). Thus, the sampling point positions and order are identical for both \( P_t \) and \( P^*_t \). This ensures more accurate loss calculation during training and minimizes additional errors caused by mismatches between pre-orthodontic and post-orthodontic point clouds. This one-to-one point correspondence also allows us to directly evaluate the error in the 6DoF parameters. It should be noted that not all teeth move during orthodontic procedures, as their positions remain reasonable and unaffected by orthodontic adjustments to surrounding teeth. For teeth whose positions remain unchanged, \( P^*_t = P_t \), implying zero rotation and translation vectors. We will flag these teeth accordingly, as described in Section \ref{subsection:Data re-organization}.

\subsection{Network Overview}
\begin{figure*}[t!]
    \centering
    \includegraphics[width=1.0\textwidth]{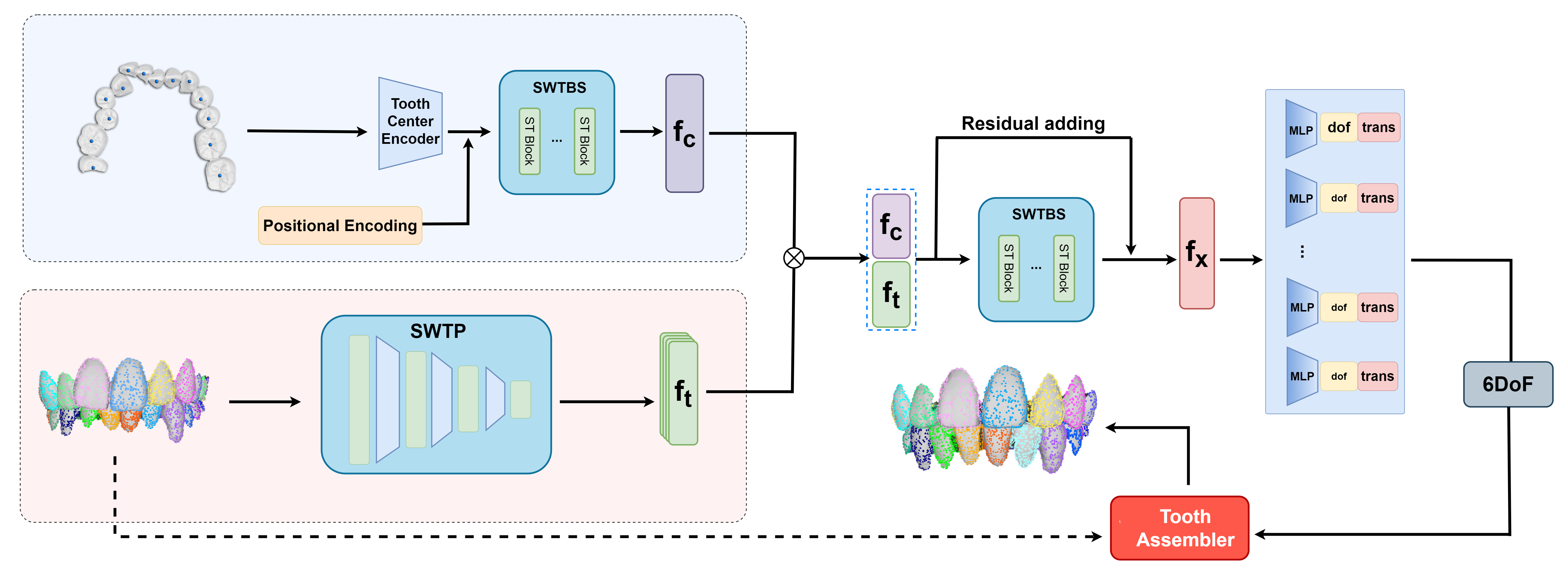}
    \caption{Our neural network architecture. The feature encoding module is divided into two branches: one encodes the global features from the tooth center, and the other encodes the local features from the tooth point cloud. The extraction of global features from the tooth center employs the SWTBS module, which consists of shared Swin-T blocks. Local features from the tooth point cloud are extracted by the SWTP module, which utilizes the multi-stage hierarchical feature fusion architecture mentioned in \cite{liu2021swin}. The hidden vectors output by the two branches are merged and then passed through SWTBS feature propagation. Finally, an MLP is used to regress the 6 degrees of freedom (6DOF) transformation parameters required for orthodontics.}
    \label{process}
\end{figure*}

The network architecture of this study, as illustrated in Figure \ref{process}, incorporates two feature extraction modules to get global and local features respectively. The global feature extraction module encodes the tooth center points using MLP layers, where the encoded latent vectors are augmented with positional encoding. These processed vectors are then passed into the Swin-T block sequence (SWTBS) to further propagate the features, with the output denoted as \( f_c \). The local feature encoder takes the three-dimensional point cloud of teeth and feeds it into the Swin transformer pipeline module (SWTP), utilizing the hierarchical downsampling module based on Swin-T's sliding window mechanism to extract features, resulting in \( f_t \).

After feature extraction, the tooth point cloud information \( f_t \) is subjected to average pooling along the tooth count dimension, then merged with the overall jaw information \( f_c \) to form a high-dimensional latent vector: \( F = \{f_c, f_t\} \). This vector is then passed through a Swin transformer block sequence module (SWTBS) for further feature extraction, yielding \( f_x \). \( f_x \) collects the residuals of \( F \) to optimize training quality. Finally, the downsampling decoder, composed of linear layers forming an MLP, regresses the 6DoF transformation parameters of teeth. These parameters, along with the initial tooth model, are passed into the Tooth Assembler module to produce the final predicted model.

The SWTBS is responsible for transmitting the central features of teeth. It is composed of Swin Transformer blocks (referred to as ST blocks) from \cite{liu2021swin}, which form one of the backbones of our network. SWTBS consists of a total of 4 connected shared Swin-T blocks, with the residuals from each block operation added to the final module output. This approach allows for the iterative amplification of high-dimensional features in the data, while also avoiding issues such as slow training convergence due to overly abstract features. The relative positional features between teeth can be shared during the transmission process. The position and orientation information of one tooth affects the localization of adjacent teeth, enhancing the overall tight connectivity of the maxillofacial structure and driving all teeth to mutually constrain each other to return to their correct positions.

\begin{figure}
    \centering
    \includegraphics[width=1.0\textwidth]{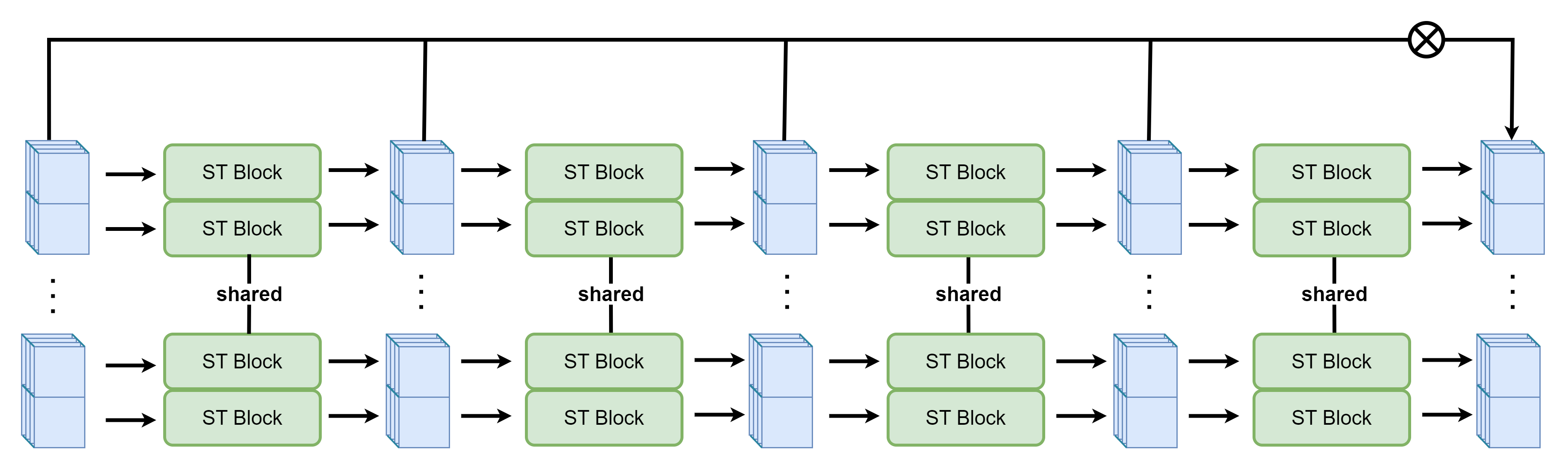}
    \caption{SWTBS module, consisting of 4 groups of shared Swin-T blocks, each group containing 16 channels. The residual of each feature transmission is added to the final output.}
    \label{swtbs}
\end{figure}

\begin{figure*}[t!]
    \centering
    \includegraphics[width=1.0\textwidth]{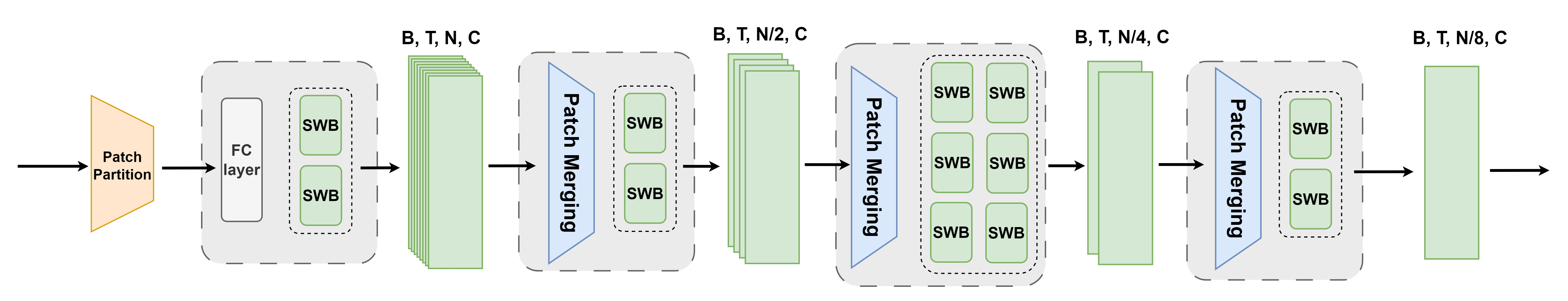}
    \caption{SWTP module, which adopts the multi-stage feature fusion mechanism from \cite{liu2021swin}, differs in that it only merges the second dimension of the latent vector and does not merge the first dimension. The first dimension represents teeth, and for orthodontic prediction, features of multiple teeth can interact but should not be fused.}
    \label{SWTP}
\end{figure*}

The tooth point cloud utilizes a hierarchical downsampling structure similar to Swin Transformer \cite{liu2021swin}, where it undergoes patch partitioning to encode into patches and then passes through four stages, each requiring feature merging and size reduction, as shown in Figure \ref{SWTP}. Unlike the default network structure mentioned in \cite{liu2021swin}, we maintain a constant channel number instead of gradually expanding the channel number. This prevents excessively high feature dimensions and potential feature loss after conversion to low dimensionality through MLP. During feature merging, we only merge data columns, which represent the number of tooth point clouds, and do not merge data rows, as height represents the number of teeth. There is no shared feature merging space between tooth counts because the rotations and translations to be predicted are specific to each individual tooth; there are no operations involving multiple teeth rotating and translating together.

\subsection{Data re-organization}
\label{subsection:Data re-organization}

\begin{figure}
    \centering
    \includegraphics[width=1.0\textwidth]{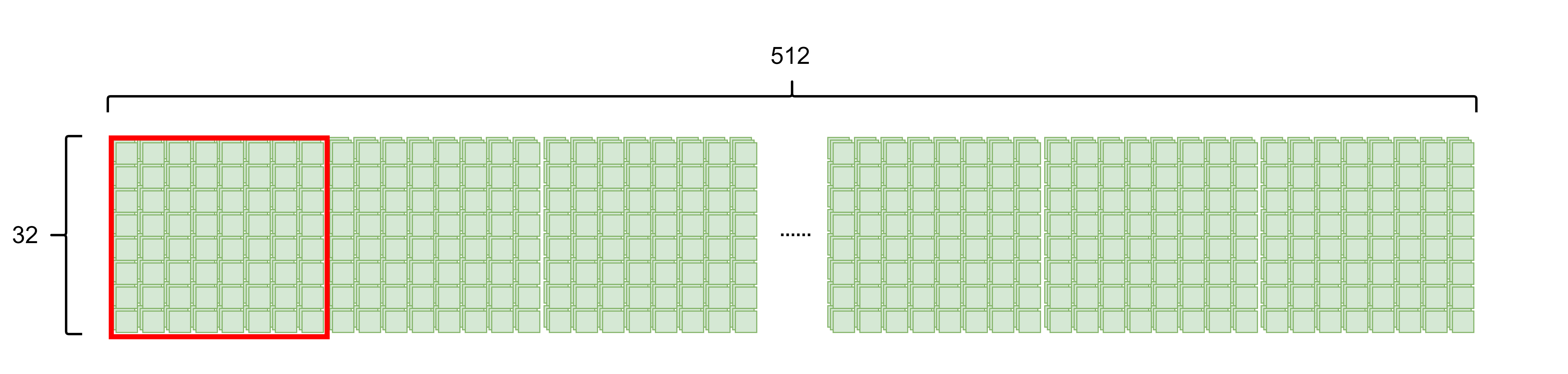}
    \caption{The organization form of a point cloud data in the dataset is Tooth Point Image. The first dimension represents 32 teeth, the second dimension represents 512 sampled points of a single tooth, and the third dimension represents the xyz coordinates of each sampled point. Therefore, the specification of a data point is [32, 512, 3].}
    \label{dataVis}
\end{figure}

The maxillary and mandibular teeth data each have a maximum of 16 teeth. Each tooth is sampled with 512 points, with each point having XYZ coordinates, forming a 3-channel matrix of size 32*512. This can be regarded as a form of color float image with dimensions of 512 in width and 32 in height, as shown in Figure \ref{dataVis}. This data organization facilitates the use of the Swin-T network structure. If the network's sampling window size is 32x32, one of them may be as depicted by the red box in Figure \ref{dataVis}. The vertical arrangement represents the order of teeth. Since the network's sampling window is close to or equal to the number of teeth, the vertical arrangement nearly encompasses all the data, and the effect of different sorting methods is minimal. This paper directly sorts the teeth based on their index numbers. The horizontal arrangement corresponds to the sorting of points within each tooth. Since the horizontal length is much larger than the network's sampling window, the horizontal order has a significant impact on the results. Considering the transmission of information between different teeth, we hope that each three-dimensional point within each network sampling window is located near similar positions across all teeth, such as simultaneously on the labial or lingual sides, or in the distal or mesial positions.

\begin{figure*}[t!]
    \centering
    \includegraphics[width=1.0\textwidth]{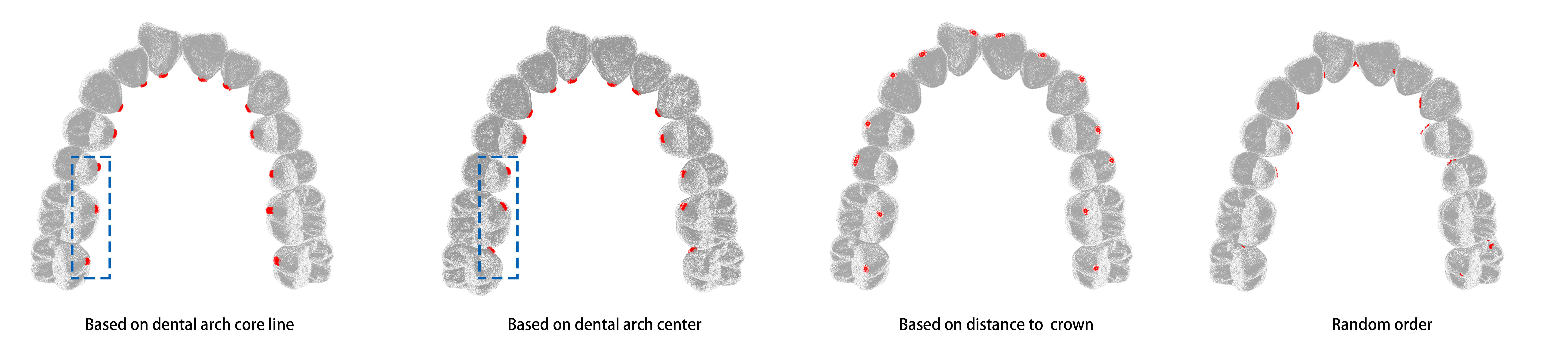}
    \caption{The distribution of points selected by the first window for each tooth after different sorting methods is visualized. The red areas indicate the locations of the first thirty points. It can be observed that the red points are most reasonably distributed when sorted based on the distance along the simulated dental arch line. In this case, all red regions are in the same local area of the tooth, thus the points selected by each window can represent the relative positions of all teeth to a certain extent.}
    \label{sortCompare}
\end{figure*}

The most straightforward sorting method is random sampling sorting. The design principle of the transformer makes it more sensitive to sequential data, and it performs excellently in extracting features from serialized data; however, this random sampling sorting method may lead to the loss of regularity information in the data, undermining the advantage of the transformer in processing data. Moreover, as shown in Figure \ref{sortCompare} a, for the red window in Figure \ref{dataVis}, the distribution of corresponding points under random sorting exhibits significant randomness, which is unfavorable for the strategy of representing the relative positions of teeth in the localized point cloud selected by the window. Another convenient sorting method is based on a specific local coordinate axis of the teeth. We adopt the local Z-axis of the teeth, so the higher points are placed in front. However, the positions of the highest points on a local coordinate axis of the teeth are irregular. For example, when sorting based on the local Z-axis, the relative positions of the localized point cloud of the posterior teeth selected by the window are not consistent with those of other teeth, as indicated by the dashed box in Figure \ref{sortCompare} b. This can result in the window failing to capture the features of the posterior teeth relative to other teeth, weakening the advantage of using the multi-level shift window structure. The third method is sorting based on the distance from the tooth's center to the gum, that is, according to the Euclidean distance from each point to the center point of the entire oral cavity. This sorting method can already achieve a relatively stable sorting, and the areas corresponding to the red window in Figure \ref{dataVis} are all located internally. However, since the distribution of teeth in the oral cavity is usually U-shaped, the method based on the distance from the tooth's center to the gum may introduce a certain angle deviation for the areas corresponding to the posterior teeth, as shown by the red dots in the blue dashed box in Figure \ref{sortCompare} c.

\begin{figure}
    \centering
    \includegraphics[width=1.0\textwidth]{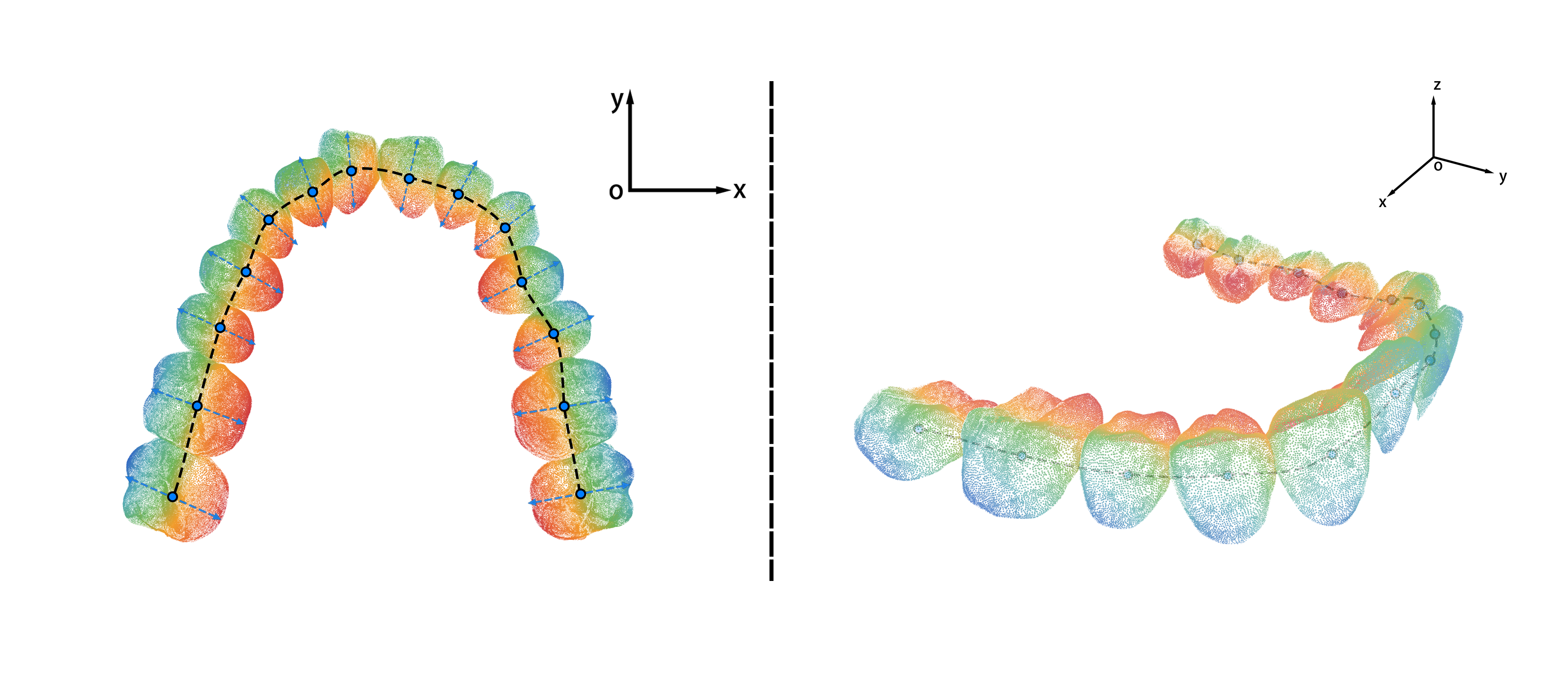}
    \caption{This figure shows the visualization of serialization. The points in the tooth point cloud are serialized according to their distance from the simulated dental arch line, with distances on the lingual side set as positive values and distances on the labial side set as negative values. Therefore, the smaller the value, the lower the color temperature, and the closer it is to blue.}
    \label{sortVis}
\end{figure}

This paper proposes a serialization method based on a simulated dental arch line. The dental arch line is a well-recognized concept in the field of dentistry, and its determination typically requires considering information about teeth, gums, and jawbones. We have designed and implemented a simulated dental arch line based on the fitting of central points from a tooth segmentation model, connecting the central points of adjacent teeth using Hermite curve interpolation. The serialization rule is to sort points by their distance from the simulated dental arch line in ascending order. Points on the labial side have positive distances from the dental arch line, while points on the lingual side have negative distances. The sorted results are shown in Figure \ref{sortVis}, where the black dashed line represents the simulated dental arch line.

Serialization based on the simulated dental arch line ensures that the 512 sampled points of each tooth maintain a consistent relative position on the tooth in a specific order. Specifically, the sampling window size is n*n, meaning that n teeth are selected at a time, with n points for each tooth. The relative position of the sub-point cloud composed of these n points represents the relative position of the entire tooth point cloud, as shown in Figure \ref{sortCompare}d. This approach allows the features learned by each point to focus more on its relative position within the red area in Figure \ref{sortCompare}d, which is more beneficial for orthodontic tasks compared to focusing on the nearby area of the point. This is also why the Swin-T block with a moving window outperforms the simple vision transformer block in the backbone network. Subsequent experiments have also demonstrated that the serialization based on the simulated dental arch line is helpful for the transformer to extract global features from the point cloud.

\subsection{Constraint Data augmentation}
\label{subsection:Constraint Data augmentation}

Existing augmentation methods are unconstrained, which can easily lead to data distortion issues, such as teeth embedding into each other or teeth shifting too far from the dental arch line. To address this, we propose a constrained data augmentation method. For both the maxilla and mandible, our augmentation process starts from the most central incisor and extends outward to the molars at both ends. For each tooth encountered, we apply random rotations based on the ground truth and random translations within a normal distribution. The randomly transformed results serve as pre-orthodontic data, while the ground truth itself remains as the corresponding post-orthodontic data, thereby generating new data-label pairs. We control the rotation noise within the range of [-10°, 10°]. The components of the translation noise vector are set according to the formula in Section \ref{subsubsection:Jaw Regularization Constraint}, with \( \mu \) being 0, and through repeated experiments, the optimal \( \rho \) is found to be half of that in Section \ref{subsubsection:Jaw Regularization Constraint} (0.3). However, such random rotations and translations often result in excessive tooth gaps, significant deviation from the simulated dental arch line, and collisions or embeddings. To address these issues, we introduce dental regularization constraints and collision detection constraints to individually optimize their positions, placing them in more reasonable pre-orthodontic positions. It is important to note that both constraints are based on the simulated dental arch line.

\subsubsection{Jaw Regularization Constraint}
\label{subsubsection:Jaw Regularization Constraint}

\begin{figure}
    \centering
    \includegraphics[width=1.0\textwidth]{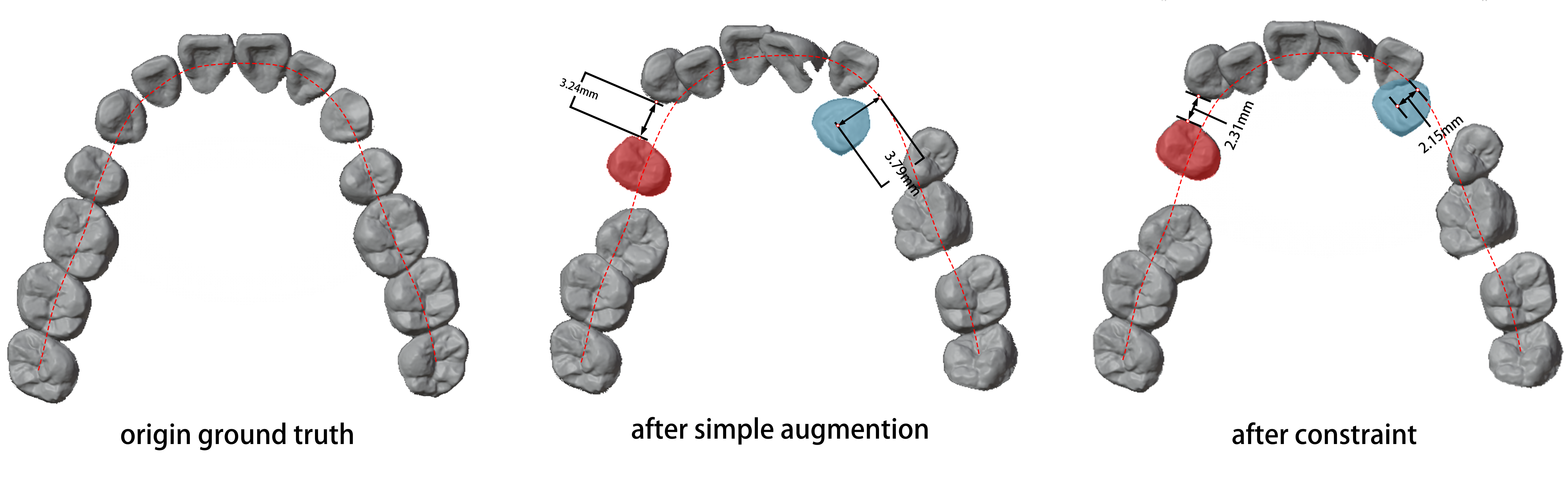}
    \caption{The process of maxillofacial regularization constraint. First, the simulated dental arch line is calculated based on the GT data; then conventional data augmentation and random transformations are performed; finally, the positions of teeth with excessively large gaps and those too far from the simulated dental arch line are corrected, resulting in the maxillofacial regularization constraint outcome.}
    \label{jaw_regular}
\end{figure}

If the distance between two teeth, excluding the missing teeth, exceeds a certain threshold \( x = 2.35 \text{ mm} \), the tooth farther from the central incisor will be moved towards the central incisor along the simulated dental arch line until the gap distance is within the threshold, as illustrated by the red teeth in Figure \ref{jaw_regular}. During the traversal process, if any tooth is excessively far from the simulated dental arch line, beyond the acceptable range for deformed teeth (as determined by dataset statistics, set to [0, 2.2 mm]), it will also be pulled inward to within the normal range, as shown by the blue teeth in Figure \ref{jaw_regular}.

\subsubsection{Collision Detection Constraints}

\begin{figure*}[t!]
    \centering
    \includegraphics[width=1.0\textwidth]{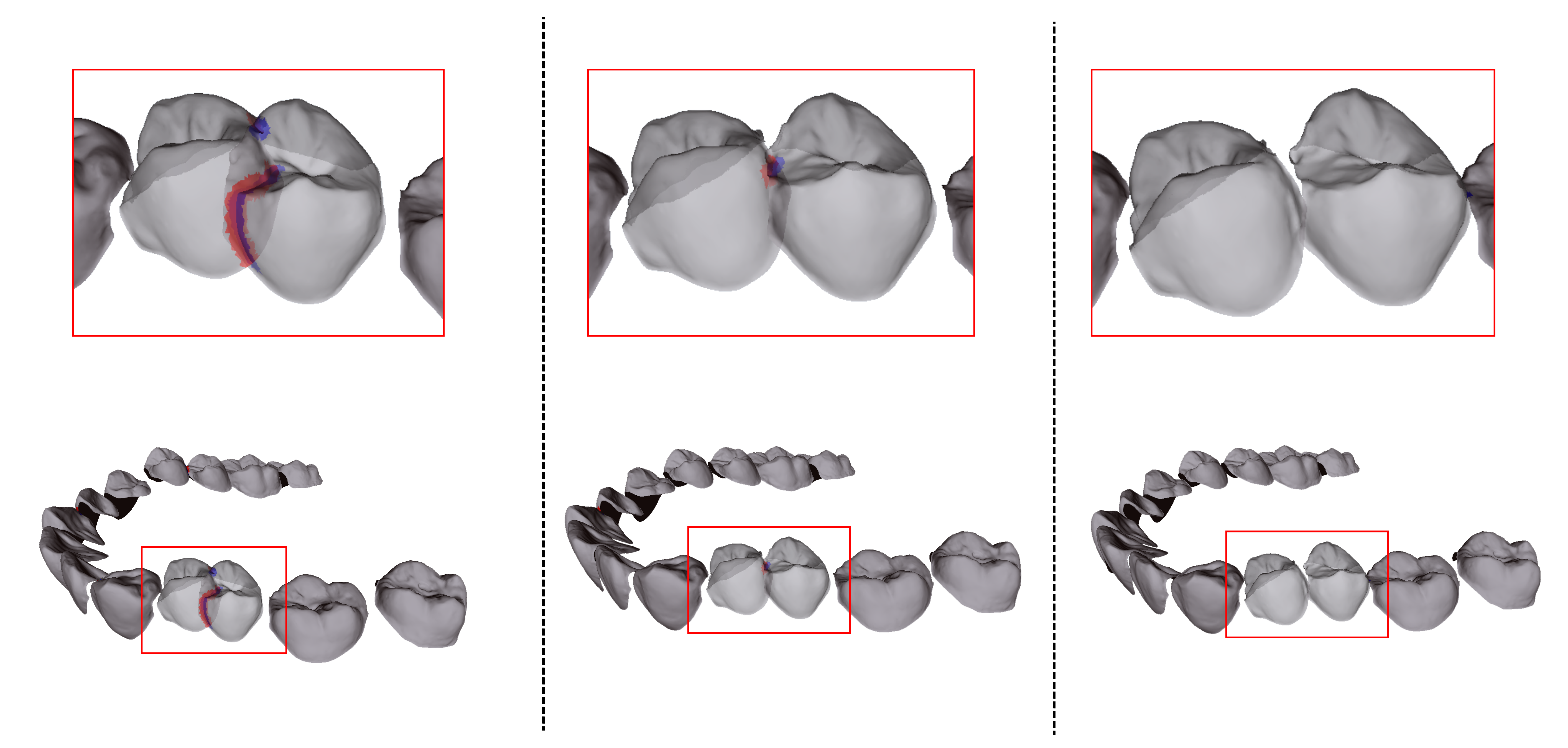}
    \caption{The iterative process of collision detection and avoidance. After detecting a collision, the teeth are displaced based on the embedding distance. Collision detection is then repeated, and if collisions still exist, displacement and detection continue until there is a proper fit with no collisions. The iterative process typically repeats 1-3 times.}
    \label{col_iter}
\end{figure*}

Non-interlocking teeth are a rigid requirement to conform to actual conditions. Therefore, we introduce an algorithm using BVH (Bounding Volume Hierarchy) collision detection\cite{gu2013efficient}\cite{pan2012fcl} to detect collisions between teeth and identify the colliding parts. Upon detecting a collision, we use the simulated dental arch line as the trajectory for collision avoidance. This approach preserves the original arch shape to some extent and keeps the augmented transformations within a normal range. Specifically, the calculation of the interlocking distance \( Dis_{col} \) between teeth \( T_{A} \) and \( T_{B} \) is defined as the farthest distance from the interlocking points of \( T_{A} \) to those of \( T_{B} \). After detecting a collision, \( Dis_{col} \) is calculated, and the tooth farther from the central incisor moves outward along the simulated dental arch line by the distance \( Dis_{col} \). This outward movement significantly reduces interlocking, but it may not completely eliminate it, as there may be a deviation between the direction of the simulated dental arch line and the line connecting the two farthest points. To ensure that the interlocking parts are fully separated, the process of collision detection and avoidance is iterated several times until no collisions are detected between any adjacent teeth. In our experiments, the number of iterations typically does not exceed three. Figure \ref{col_iter} shows the iterative process of collision avoidance.

\subsection{Loss function}

The global loss function of the network consists of four components: model reconstruction loss, rotation and translation parameter loss, occlusion projection range consistency loss, and occlusal distance uniformity loss. The first two losses are employed in existing methods, while the latter two are specifically designed in this paper to address dental occlusion, a critical medical indicator. These losses aim to ensure that the predicted results meet the requirements for dental occlusion. The detailed calculation methods for these losses will be elaborated in subsequent sections. Given the differences in magnitude and importance among the loss function components, each part will be weighted by corresponding hyperparameters for adjustment.

\begin{equation}
    L = \delta_{0}*L_{recon} + \delta_{1}*L_{fit} + \delta_{2}*L_{uni} + \delta_{3}*L_{val}
\end{equation}

The overall loss \( L \) consists of:
- \( L_{\text{recon}} \): model reconstruction loss,
- \( L_{\text{fit}} \): occlusal projecting overlap loss,
- \( L_{\text{uni}} \): jaw occlusal distance uniformity loss,
- \( L_{\text{val}} \): transformation parameters loss (translation and rotation).
The hyperparameters \( \delta_0 \) through \( \delta_3 \) are used to weight each component accordingly.

\subsubsection{Reconstruction Loss}

We utilized the model reconstruction loss mentioned in \cite{wang2022tooth}, calculating the point-by-point discrepancies between the predicted results and the ground truth as the loss. This approach effectively supervises the gap between the predictions and the ground truth, significantly aiding the prediction of model translation and rotation. Additionally, the reconstruction loss for each tooth includes a loss component for the deviation of the predicted tooth center \( c_t \) from the ground truth. Unlike \cite{wang2022tooth}, the post-orthodontic data in our dataset is manually adjusted by orthodontists using pre-orthodontic data. Consequently, the vertex positions and orders of the pre- and post-orthodontic models correspond one-to-one, allowing the reconstruction loss to be obtained by subtracting the corresponding vertices of the pre- and post-orthodontic models. If post-orthodontic scanned models were used as post-orthodontic data, vertex correspondence would be based on the nearest distance, which could lead to calculation errors when the teeth are too far apart to correctly match the points.

\begin{equation}
	L^{point}_{recon} = \sum_{t \in T} \left ( \sum_{i = 0, p \in P_t} || \overline{p}_i - p^{*}_i || _2^2 + || \overline{c}_t - c^{*}_t || _2^2 \right )
\end{equation}

Here, \( T \) represents the set of all teeth, \( t \) represents an individual tooth within this set, \( P \) denotes the point cloud information of the teeth, and \( \overline{p} \) and \( \overline{c} \) respectively denote the initial tooth model point cloud and initial tooth center position transformed using predicted orthodontic transformations. \( p^* \) and \( c^* \) represent the corresponding ground truth point cloud model and tooth center position.

\subsubsection{Rotation\&translation numerical loss}

The transformation parameter loss comprises two components: rotation loss and translation loss, calculated as the L1 loss between the quaternion and translation vector predictions and ground truth \cite{huang2022github}. Given the significant misalignment often present in the original teeth, correcting misaligned teeth to their aligned positions is a crucial task of this network. Therefore, this study computes rotation enhancement weights \( \zeta^{\text{rotate}}_t \) and translation enhancement weights \( \zeta^{\text{trans}}_t \) for each tooth during training, based on the magnitude of misalignment. These weights are cumulatively added to the original loss, as shown in Equation \ref{rot_loss_eq}/\ref{tran_loss_eq}. Drawing from \cite{huang2022github}, we emphasize that more severe misalignments should receive greater attention, corresponding to larger loss values. \( \zeta^{\text{rotate}}_t \) normalizes the rotation angle of each tooth from pre- to post-orthodontic using \( \frac{\pi}{2} \) (with a maximum limit of 1.0 if the rotation angle exceeds \( \frac{\pi}{2} \)); \( \zeta^{\text{trans}}_t \) normalizes the translation vector magnitude using \( \text{Max}_t \) (the maximum translation distance limit statistically observed in the dataset, set at 4.5). Both weights range within [0, 1]; during testing, where ground truth is unavailable, both are set to 1.0. Throughout training, teeth models with larger transformation magnitudes from initial to ground truth positions have higher weights \( \zeta \), emphasizing areas needing optimization and improving network learning quality.

Compared to model reconstruction loss, which computes the discrepancy between predicted and ground truth at the result level, rotation and translation parameter losses measure the numerical disparities in transformation parameters towards the final outcome. These losses can represent the degree of similarity between predicted results and ground truth at higher-order dimensions. When combined with model reconstruction loss, they further emphasize the role of ground truth labels.

\begin{equation}
    \label{rot_loss_eq}
	L_{rotate} = \sum_{t \in T} \left[ \sum_{i = 0}^{3} L_{1} \left( \overline{rotate}_{t}(i), rotate^{*}_{t}(i) \right) * \left( 1.0 + \zeta^{rotate}_{t} \right) \right]
\end{equation}

\begin{equation}
    \label{tran_loss_eq}
	L_{trans} = \sum_{t \in T} \left[ \sum_{i = 0}^{2} L_{1} \left( \overline{trans}_{t}(i), trans^{*}_{t}(i) \right) * \left( 1.0 + \zeta^{trans}_{t} \right) \right]
\end{equation}

\( \overline{rotate}_t \), \( \overline{trans}_t \), \( \text{rotate}^*_t \), and \( \text{trans}^*_t \) respectively denote the predicted rotation quaternion and translation vector, as well as the ground truth rotation and translation for tooth \( t \). \( \zeta^{\text{rotate}}_t \) and \( \zeta^{\text{trans}}_t \) represent the rotation enhancement weight and translation enhancement weight, respectively.

\begin{equation}
	L_{val} = \omega * L_{rotate} + L_{trans}
\end{equation}

The final transformation parameter loss combines the above components. Due to the relatively small numerical values of rotation loss, an additional parameter \( w \) is introduced to amplify the impact of quaternion rotation loss during actual training. In all experiments conducted in this study, \( w \) is set to 10.0.

\subsubsection{Occlusal projecting overlap}
\label{subsubsection:Occlusal projecting overlap}

Occlusion projection range consistency loss represents whether the interocclusal region between the predicted results of upper and lower jaws matches the ground truth. The definition of occlusion projection range is as follows: Let tooth \( t \) in one jaw have a corresponding area \( \beta_t \) in the opposite jaw. Project all points of \( t \) and \( \beta_t \) onto the occlusal plane. If the closest distance between points from \( t \)'s point cloud and \( \beta \)'s point cloud (on the occlusal plane) is less than a threshold \( \tau \), then points from \( t \)'s point cloud within this distance are considered part of \( t \)'s occlusion projection range. Typically, this study sets the threshold \( \tau \) to 0.07 mm. The detailed definition of upper and lower jaw dental occlusion relationships is outlined in \cite{andrews1972six}, while the post-orthodontic data annotated by orthodontists in the dataset adhere to the six-elements standard. Therefore, we introduce the concept of occlusion projection range to bring predicted results closer to the ground truth at the occlusion projection range level, aiming to achieve correct upper and lower jaw dental correspondences that meet the requirements of the six-elements standard.

\begin{figure}
    \centering
    \includegraphics[width=1.0\textwidth]{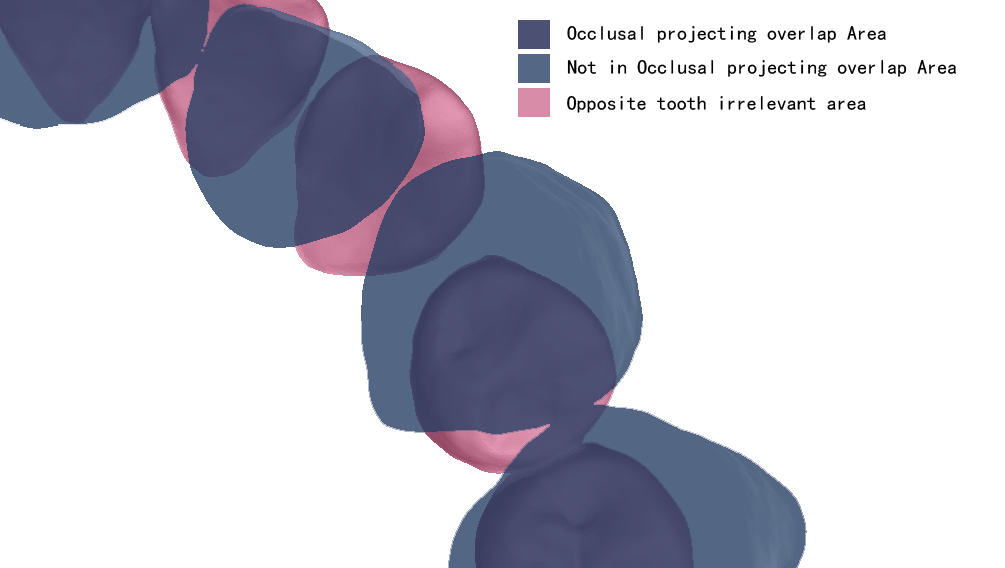}
    \caption{Visualization of the occlusion projection range. The blue area represents the projection of the upper teeth, the red area represents the projection of the lower teeth, and their overlapping region (dark purple) indicates the occlusion projection range.}
    \label{occ-fit}
\end{figure}

\begin{equation}
	m_i = \underset{p_{j} \in P^{f}_{\beta_t}}{Argmin} \left \| p_{i} - p_{j} \right \|_{2}, p_i \in P^f_t
\end{equation}

In the Figure \ref{occ-fit}, \( P_t^f \) represents the point cloud of tooth \( t \) projected onto the occlusal plane, \( P_{\beta_t}^f \) represents the point cloud of region \( \beta_t \) projected onto the occlusal plane, and \( m_i \) denotes the minimum two-dimensional plane distance between a point \( p_i \) from the point cloud of \( t \) and the nearest point \( p_j \) from the point cloud of \( \beta_t \).

\begin{equation}
	X = \left \{ \begin{aligned}
		1 (m_i < \tau),\\
		0 (m_i \geq \tau).\\
	\end{aligned}
	\right .
\end{equation}

\( \tau \) is the threshold used to divide the occlusion projection range. \( X \) is a binary sequence where \( X_t(i) \) records whether point \( p_i \) from the point cloud of tooth \( t \) belongs to the occlusion projection range based on the relationship between \( m_i \) and \( \tau \). We compare whether the occlusion projection ranges match between the predicted results and ground truth, using the degree of match as the occlusion projection range consistency loss function:

\begin{equation}
	L_{fit} = \underset{t \in T}{Ave} \left ( \sum_{i=0}^{n-1} \left | \overline{X}_t(i) - X^{*}_t(i) \right | \right )
\end{equation}

Finally, \( L_{\text{fit}} \) represents the result of the occlusion projection range consistency loss. \( \overline{X}_t(i) \) denotes the predicted label for the \( i \)-th point of tooth \( t \), while \( X^*_t(i) \) represents the ground truth label for the \( i \)-th point of tooth \( t \). \( n \) represents the number of elements in \( \overline{X}_t \).

\subsubsection{Occlusal distance uniformity}

Due to the completely different morphologies and occlusion patterns between anterior and posterior teeth, we have proposed distinct loss function designs for anterior and posterior teeth based on discussions in \cite{davies2001occlusion}, \cite{hiew2006optimal}, and \cite{andrews1972six}. Therefore, the occlusal distance uniformity loss \( L_{po} \) across upper and lower jaws is composed of two parts: \( L_{po}^i \) for anterior teeth and \( L_{po}^g \) for posterior teeth. As shown in the following formula, we introduce a weighting parameter \( w^g \) for the posterior teeth to balance their importance. By default, \( w^g \) is set to 2.0.

\begin{equation}
    L_{uni} = L^{ant}_{uni} + w^{pior} \cdot L^{pior}_{uni}
\end{equation}

\textbf{Posterior jaw occlusal distance uniformity}

\begin{figure}
    \centering
    \subfigure{
        \includegraphics[width=0.45\textwidth]{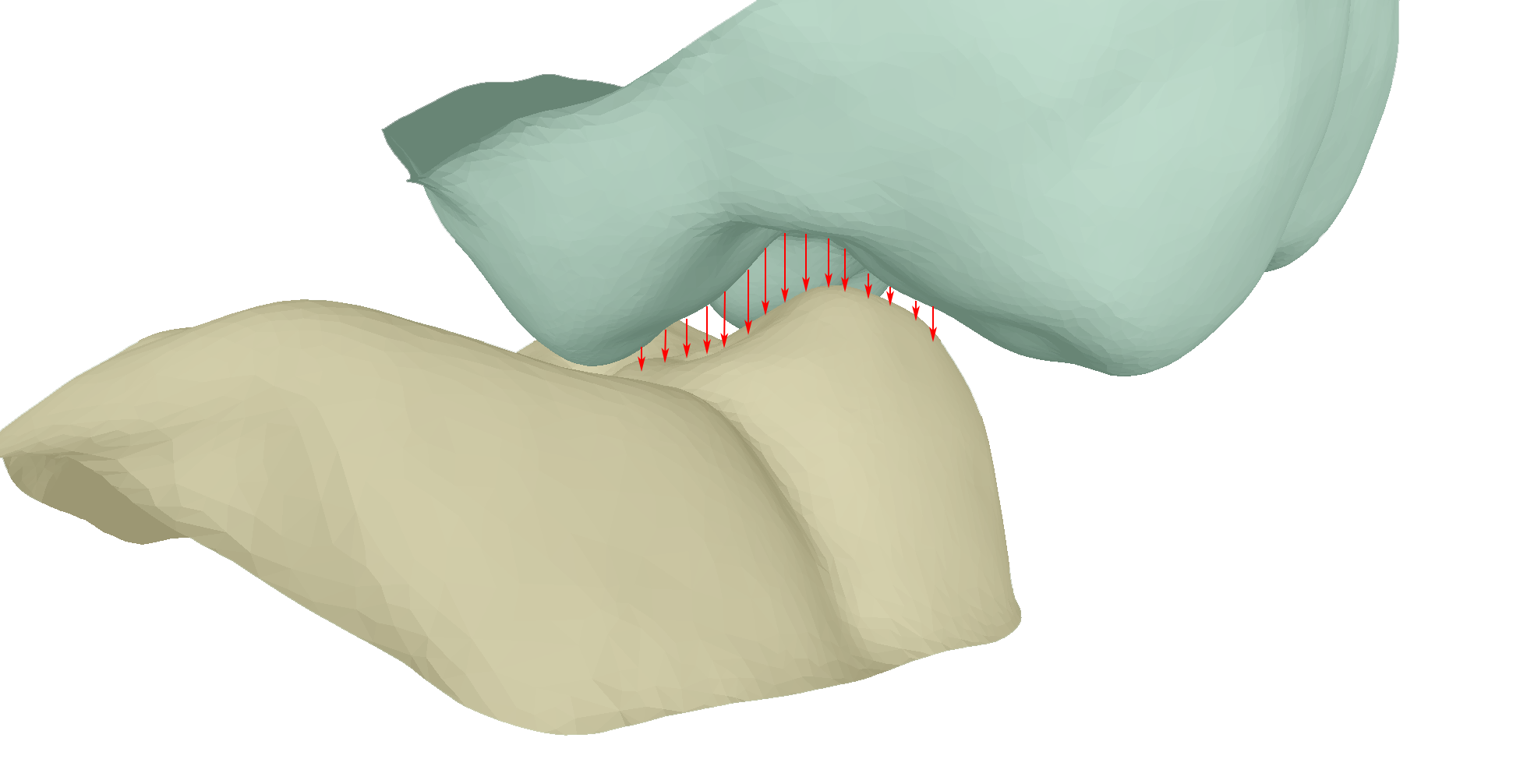}
        \label{occ-dis-uni-1}
    }
    \subfigure{
        \includegraphics[width=0.45\textwidth]{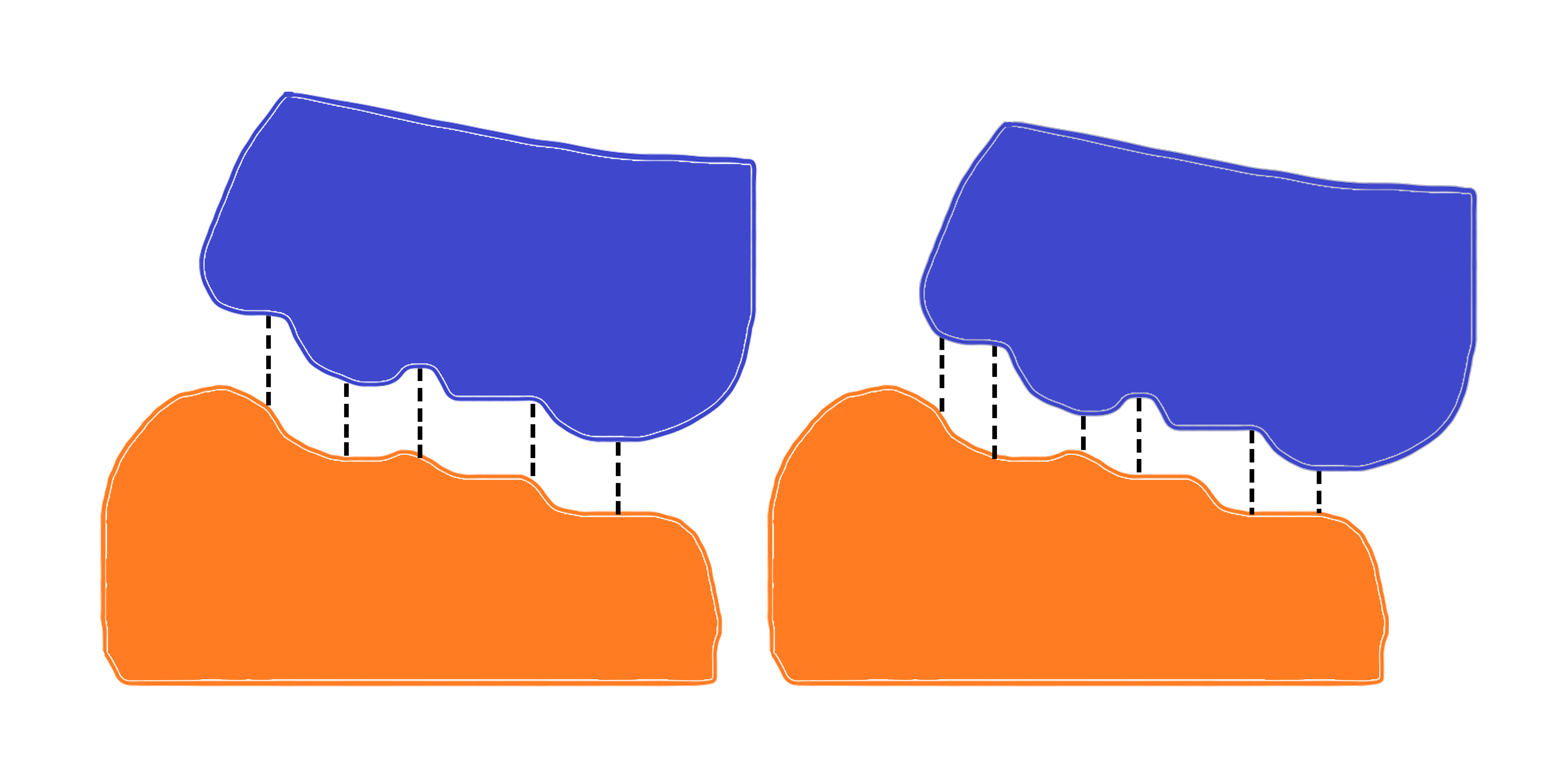}
        \label{occ-dis-uni-2}
    }
    \caption{Visualization of occlusion distance. The length of the red line segments within the occlusion projection range in the left image represents the occlusion distance. It can be observed that the occlusion distance on the left side of the right image is much more uniform compared to the right side, indicating that the uniformity loss of the occlusion distance on the left side of the right image is smaller.}
    \label{occ-dis-uni}
\end{figure}

In the constraint of projection ambiguity discussed in \cite{davies2001occlusion}, it is mentioned that the direction and magnitude of vectors connecting corresponding points in the occlusal regions of slot teeth can assess the reasonableness of occlusion. Therefore, this paper focuses on evaluating the consistency and similarity of connecting vectors between corresponding points in the occlusal regions of posterior teeth. The occlusal distance uniformity of posterior teeth is calculated within the occlusal projection range defined in section \ref{subsubsection:Occlusal projecting overlap}. For a point \( p_i \) in the occlusal projection range of tooth \( t \), the distance to the nearest point \( p_j \) in the corresponding occlusal projection range \( \beta_t \) on the opposing jaw is denoted as \( d \). The collection of all such distances \( d \) from tooth \( t \) forms set \( D \). The uniformity of \( D \) represents the degree to which the occlusal ranges of upper and lower teeth match concavely or convexly, thereby defining the occlusal distance uniformity loss function for posterior teeth:

\begin{equation}
	L^{pior}_{uni} = \sum_{t \in T_{pior}} \underset{X_t(i) = 1}{Var} \left( \min_{X_{\beta_t}(j) = 1} \left \| p_{i} - p_{j} \right \|_{2} \right)
\end{equation}

\( X_t(i) \) and \( X_{\beta_t}(j) \) both equal to 1.0 represent that only points within the occlusal projection range of tooth \( t \) and its corresponding tooth region \( \beta_t \) are considered. \( L^g_{po} \) denotes the result of the occlusal distance uniformity loss function for posterior teeth.

\textbf{Anterior jaw occlusal distance uniformity}

\begin{figure}
    \centering
    \includegraphics[width=0.8\textwidth]{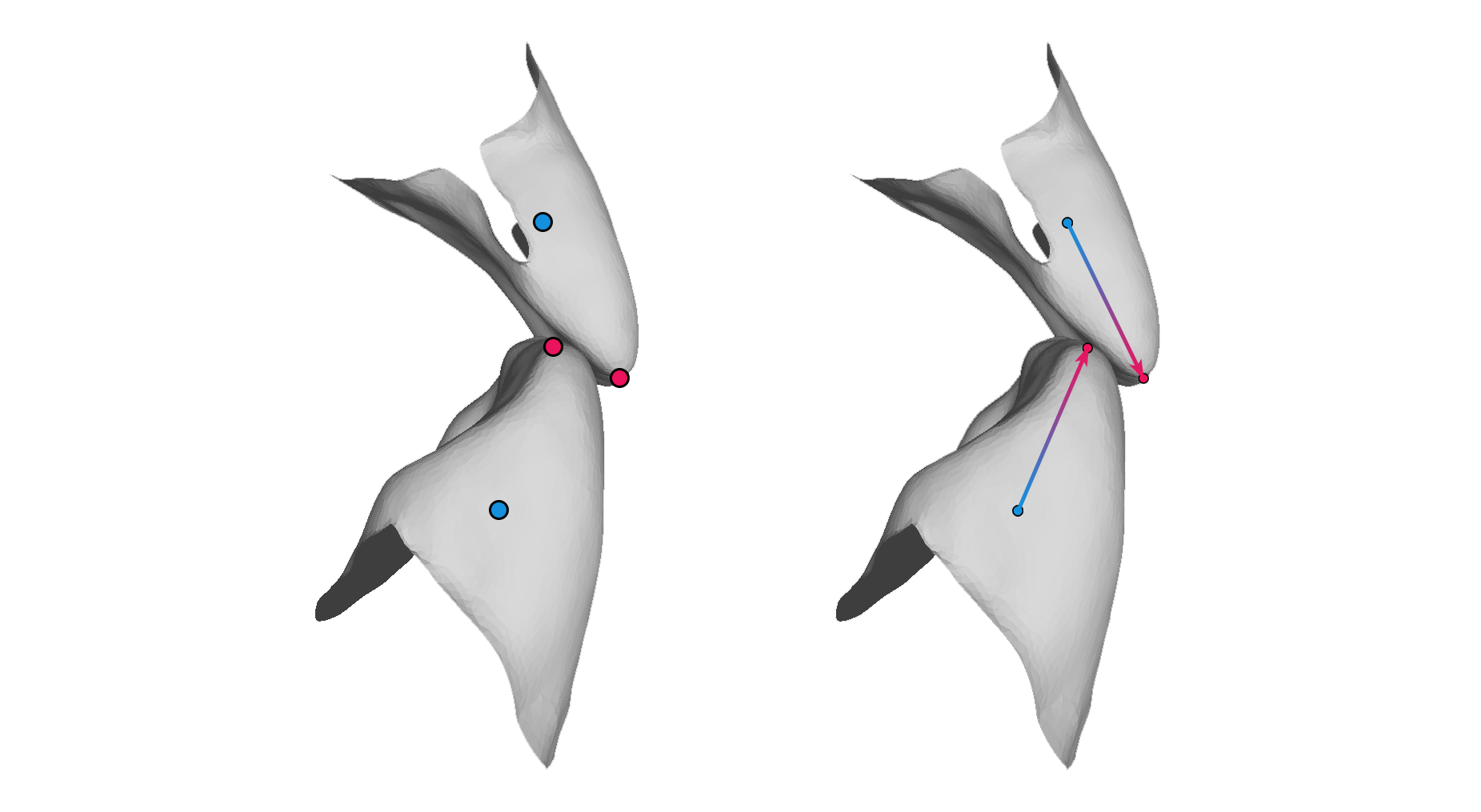}
    \caption{The structure of the anterior teeth is relatively special, so the uniformity of occlusion distance is calculated based on the tooth centers and crown points. This is assessed by comparing the differences between the predicted data and GT data for the central points and crown points, as well as the angular differences of the vectors pointing from the central points to the crown points.}
    \label{ant-occ-dis-uni}
\end{figure}

The third element of the six elements\cite{andrews1972six}, as mentioned in dental torque occlusion, states that the upper anterior teeth should maintain a certain degree of inclination and be positioned anteriorly to the lower anterior teeth. It is also noted during central occlusion \cite{hiew2006optimal} that the upper anterior teeth should correspond individually to the lower anterior teeth. The occlusal relationship of anterior teeth involves the lower teeth being appropriately enveloped by the lingual aspect of the upper teeth, with the upper anterior teeth positioned buccally and the lower anterior teeth lingually \cite{davies2001occlusion}. This perfect relative positioning is encompassed in the ground truth, as illustrated in the Figure \ref{ant-occ-dis-uni}.

The loss function for perfect anterior tooth occlusion consists of two parts, \(L^{ant1}_{uni}\) and \(L^{ant2}_{uni}\), which respectively represent the differences in vertex coordinates and angular disparities between the predicted tooth axis vectors and the ground truth. The parameter \( \omega^{ant} \) serves as a hyperparameter that assists in normalizing the data for \(L^{ant1}_{uni}\) and \(L^{ant2}_{uni}\), compensating for their magnitude differences.

\begin{equation}
	L^{ant}_{uni} = L^{ant1}_{uni} + \omega^{ant} * L^{ant2}_{uni}
\end{equation}

Let the vector from the centroid of the anterior teeth coordinates to the incisal edge be denoted as the tooth axis vector \( \mathbf{V}_c \). We calculate the angular disparity between the predicted vector \( \overline{V}_c \) and the ground truth vector \( \mathbf{V}^*_c \), as well as the positional errors at the start and end points of \( \overline{V}_c \) and \( \mathbf{V}^*_c \). These metrics serve to represent the uniformity of anterior teeth occlusion distances.

\begin{equation}
	L^{ant1}_{uni} = \sum_{t \in T_{ant}} \left \| \overline{c}_{t} - c^{*}_{t} \right \|_{2} + \sum_{t \in T_{ant}} \left \| \overline{Peak}_{t} - Peak^{*}_{t} \right \|_{2}
\end{equation}

\( T_{\text{ant}} \) represents the subset of teeth in the set \( T \) that belong to the anterior teeth. \( \overline{c}_{t} \) denotes the centroid of the predicted point cloud, and \( c^{*}_{t} \) represents the centroid of the ground truth point cloud. "Peak" denotes the highest incisal point in the point cloud collection \( t \) in the z-direction, where \( \overline{Peak}_{t} \) and \( Peak^{*}_{t} \) correspond to the highest incisal points in the predicted and ground truth data, respectively. \( L^{ant1}_{uni} \) is the first component of the loss function for perfect anterior teeth occlusion, which measures point deviation.

\begin{equation}
	L^{ant2}_{uni} = \sum_{t \in T_{ant}} arccos \left( \frac{\left( \overline{Peak}_{t} - \overline{c}_{t} \right) \cdot \left( Peak^{*}_{t} - c^{*}_{t} \right) }{\left \| \overline{Peak}_{t} - \overline{c}_{t} \right \|_{2} * \left \| Peak^{*}_{t} - c^{*}_{t} \right \|_{2}} \right)
\end{equation}

\(L^{ant2}_{uni}\) is the second component of the loss function for perfect anterior teeth occlusion, representing angular disparity.

\section{Experiments}

\subsection{Dataset and pre-processing}
Our dataset comprises 855 sets of dental data derived from 3D models of upper and lower jaw teeth constructed through oral scans. The initial segmentation is obtained using the semantic segmentation network TSegNet \cite{cui2021tsegnet}, followed by manual mesh and edge optimization. The optimized crown models are then aligned and arranged by experienced orthodontists to obtain the corresponding ground truth data. For the dataset usage, we randomly selected 755 samples for training, and selected 30 samples for validating, with the remaining 70 samples used for testing. These samples, all intended for orthodontic purposes at dental hospitals, exhibit various degrees of misalignment. Approximately 50$\%$ of the samples are severely deformed, featuring issues such as wisdom teeth, severe dental arch misalignment, missing teeth, and significant gaps between teeth. To ensure the comprehensiveness of the testing experiments, we exchange some severely deformed samples from the training set with non-severely deformed samples in the testing set.

It is important to note that our labeled data does not consist of post-orthodontic intraoral scans, as this would introduce discrepancies in the scale, sequence, and topology of the point clouds before and after orthodontic treatment. Instead, orthodontists directly apply positional transformations to the segmented tooth models from the intraoral scans to obtain the ideal post-treatment state, which serves as the labeled data. Compared to previous methods, this approach avoids the aforementioned discrepancies, making the network's predicted results closer to the ideal treatment outcome and more beneficial for assisting orthodontists in designing patient treatment plans. If the final treatment is successful, the patient's dental morphology will closely match the initial arrangement. Conversely, if the treatment is less successful, the two will not align, and such data is not a reasonable representation of dental arrangement, thereby hindering training. In actual medical cases, a considerable proportion of patients do not achieve the expected orthodontic results, rendering their actual dental morphology unusable. However, regardless of the success of the final orthodontic treatment, the morphology designed by the orthodontist is always usable. Therefore, another advantage of using orthodontist-designed arrangement data is that it allows us to fully utilize almost all orthodontic data.

\subsection{Implementation details and performance}
\label{subsection:Implementation details and performance}

We utilized an NVIDIA GeForce RTX 3090 with 24GB of VRAM for training, conducting the training over 500 epochs with a batch size of 8 data cases. To ensure the rationality and regularity of window movement, we set the number of sampling points \( N \) to 512. The initial learning rate was set to 1.5e-4, and the coefficient \( w \) for the rotation loss in the transformation parameter loss was set to 10.0 to balance the magnitudes of rotation and translation losses. The threshold \( \tau \) for determining whether the corresponding occlusal surface vertex falls within the occlusal projection range was empirically set to 0.07mm, which yielded the most appropriate occlusal projection range. The shift window size was set to 8\(\times\)8. Given that a single dataset can contain a maximum of 32 teeth, the window size was chosen to be a divisor of the maximum number of teeth.

\subsection{Evaluation metrics}
Our evaluation metrics employed the Add/AUC as proposed in the TANet \cite{tanet2020} and landmark \cite{wang2022tooth} papers. AUC represents the area under the add curve when \( k = 5 \), where the add curve describes the point-to-point error of the point cloud model. The x-axis variable represents the maximum error threshold, and the y-axis variable represents the proportion of points in the point cloud model with errors less than x. The value \( k \) denotes the maximum value of the x-axis variable. The value of ADD/AUC is the ratio of the integral of the add curve over the interval [0, k] to \( k \). Similarly to landmark \cite{wang2022tooth}, we also calculated the average rotation error and translation error. The data presented in the table \ref{four_exp} records the results of four testing experiments conducted using our network.

\begin{table}
	\caption{Records of results from 4 experiments, with identical conditions for each, indicating that the network's experimental results are stable.}
	\centering
	\begin{tabular}{llll}
		\toprule
		Exp id & ADD/AUC & $ME_{rotate}$ & $ME_{translate}$ \\
		\midrule
		exp 1 & 0.89 & 2.5314  & 1.0782 \\
		exp 2 & 0.89 & 2.7678  & 1.1584 \\
		exp 3 & \textbf{0.90} & \textbf{2.7513}  & \textbf{1.0573} \\
        exp 4 & \textbf{0.92} & \textbf{2.3254}  & \textbf{1.0167} \\
		\bottomrule
	\end{tabular}
	\label{four_exp}
\end{table}

\subsection{Comparisons with STAT methods}

\subsubsection{Comparison of indicators}

We conducted training and testing using the methods TANet \cite{tanet2020}, PSTN \cite{PSTN2020}, TAligNet \cite{iorthopredictor2020}, and Landmark \cite{wang2022tooth}. TANet, PSTN, and TAligNet were reproduced according to the detailed network structures, parameters, and loss functions described in the corresponding papers. The FPM module of TANet utilized the source code provided in \cite{huang2022github}. The results for the landmark method were obtained by the original authors using the data we provided. During the reproduction process, parameters such as the number of channels, data augmentation settings, and the number of sampling points were configured according to the recommended values in the papers. The training and testing data, as well as the training specifications for these methods, were consistent with those described in Section \ref{subsection:Implementation details and performance} of this paper. We compared the AUC, average rotation error, and average translation error of these methods. As shown in Table \ref{stat_compare}, our method outperforms the others in all aspects. Whether in terms of AUC, reflecting the similarity to the Ground Truth, or the rotational and translational deviations in the aligned state, our method achieved superior results.

\begin{table}
	\caption{Comparison of evaluation metrics between the proposed method and the STAT method.}
	\centering
	\begin{tabular}{lllll}
		\toprule
		\multirow{2}{*}{Model} & \multicolumn{4}{c}{Test result} \\
		\cline{2-5}
		 & ADD $\downarrow$  & ADD/AUC $\uparrow$ & $ME_{rotate} \downarrow$ & $ME_{translate} \downarrow$ \\
		\midrule
		TAligNet       & 1.5307 & 0.72 & 7.5461 & 2.0392 \\
		TANet          & 1.0075 & 0.81 & 6.9274 & 1.6815 \\
		PSTN           & 1.5889 & 0.71 & 8.6938 & 2.2155 \\
		Ptv3           & 1.2136 & 0.78 & 7.0663 & 1.7581 \\
		$Landmark^{*}$ & 0.7972 & 0.84 & 7.2450 & 1.3457 \\
		$Landmark$     & 0.8139 & 0.84 & 7.8277 & 1.3764 \\
		Ours           & \textbf{0.6584} & \textbf{0.89} & \textbf{2.7678} & \textbf{1.1584} \\
		\bottomrule
	\end{tabular}
	\label{stat_compare}
\end{table}

\begin{figure}
    \centering
    \includegraphics[width=1.0\textwidth]{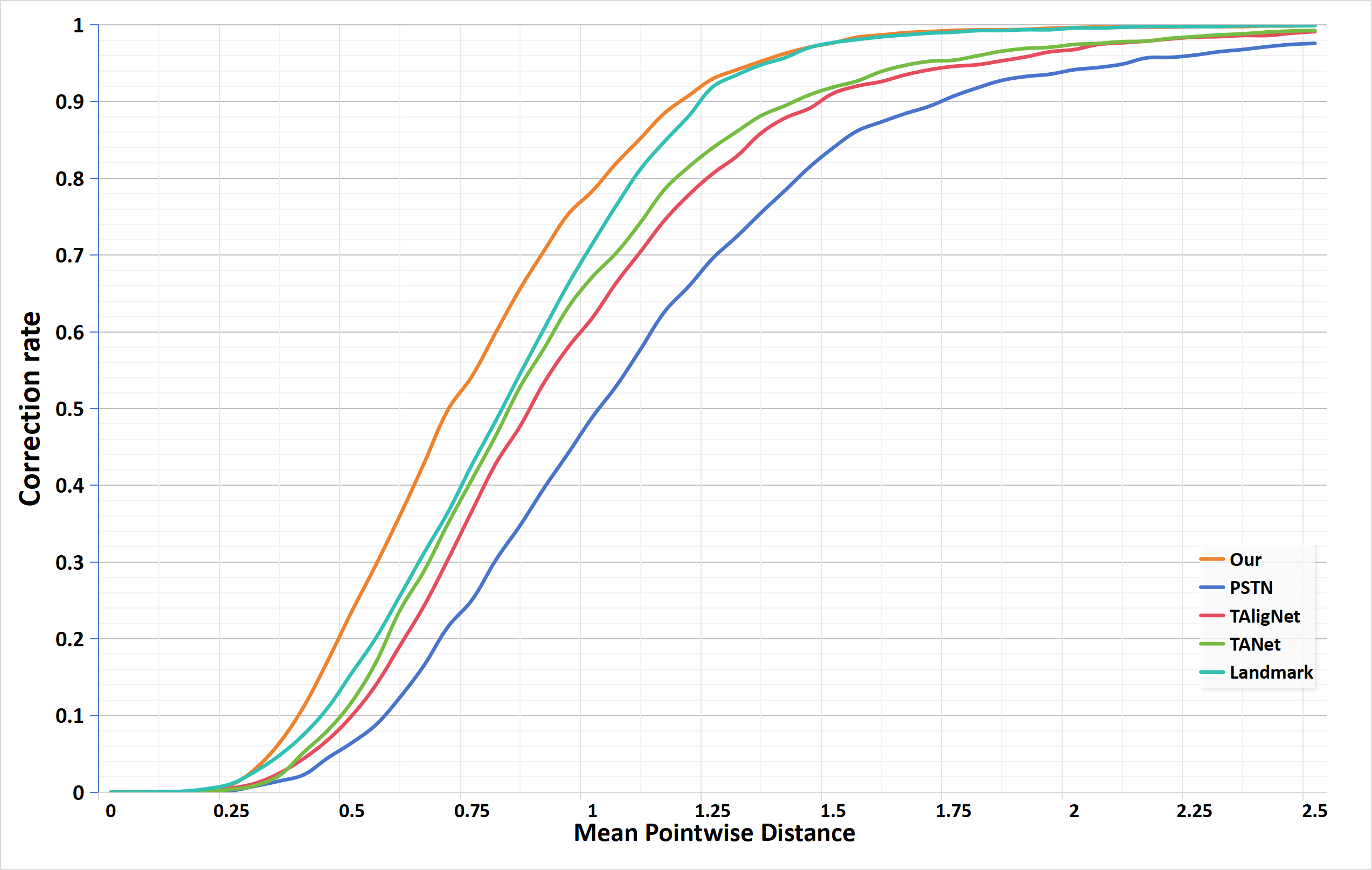}
    \caption{Comparison of accuracy curves between the proposed method and the STAT method (under different average point distances).}
    \label{auc_list}
\end{figure}

We compared the curves of average point distances, as shown in Figure \ref{auc_list}. It is evident from the figure that our method achieves the highest accuracy under different definitions of average point distance. It is noteworthy that beyond an average point distance of 2.5, all curves converge to nearly 1.0. Therefore, the chart only displays curves for \( k \leq 2.5 \).

\subsubsection{Comparison of Result Quality}

\begin{figure}
    \centering
    \includegraphics[width=1.0\textwidth]{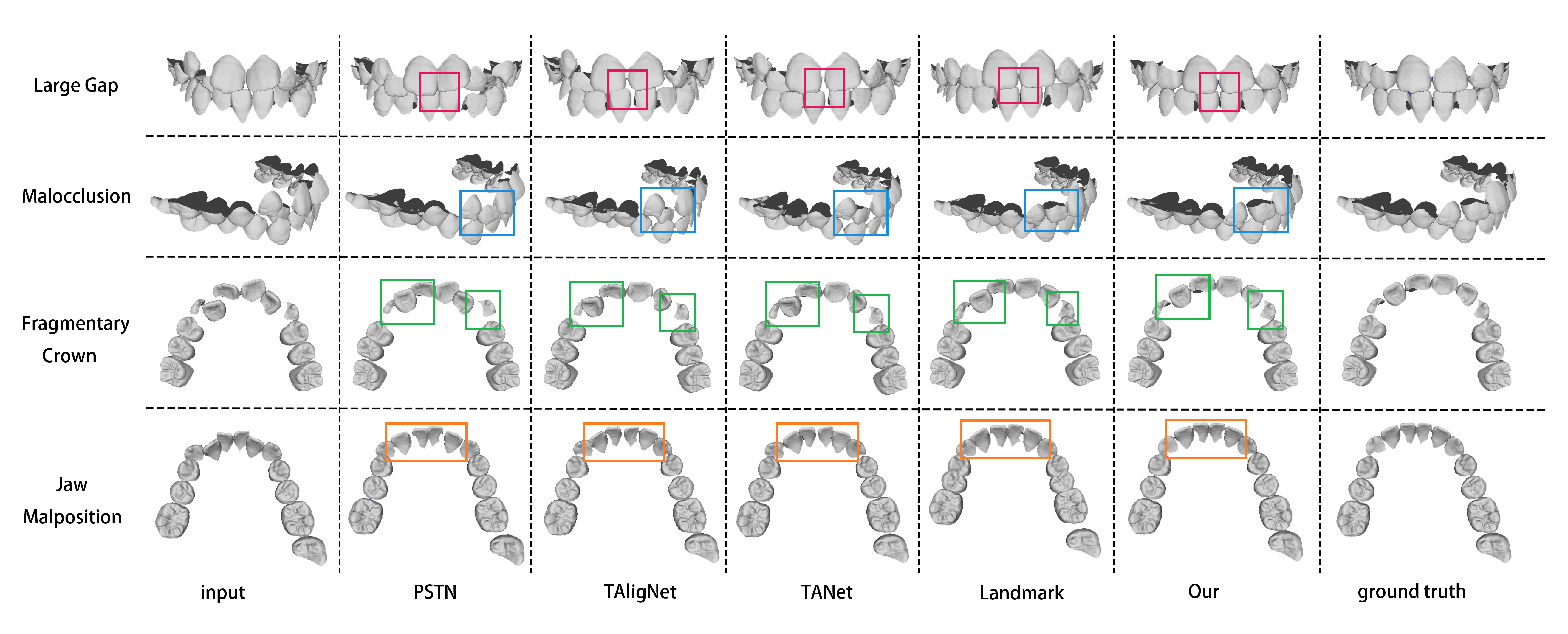}
    \caption{Comparison of prediction results between the proposed method and other methods. Here, four typical and challenging orthodontic problems are selected, listed from top to bottom: large gap, malocclusion, fragmentary crown, and jaw malposition.}
    \label{quality_compares}
\end{figure}

Figure \ref{quality_compares} compares the aligned tooth models obtained by applying the transformation parameters predicted by our method and other methods to the initial model. The comparisons are recorded from the frontal view, side view, and top view of the upper and lower jaws. The first two rows highlight the issues of misaligned upper and lower front teeth gaps and the occlusal misalignment of the upper and lower jaws. During training, our occlusal projection range alignment loss restricts the relative positions of the upper and lower teeth to the correct positions, and the occlusal distance uniformity loss constrains the occlusal points of the upper and lower crowns to be closer together. This effectively addresses these two issues. Compared to the Chamfer vector loss used by other methods to constrain the relative positions of the upper and lower teeth, our approach imposes more detailed requirements on the occlusal state. The introduction of landmarks makes the network more focused on the specific joint points, resulting in a good resolution of the misaligned teeth gaps.

The last two rows of Figure \ref{quality_compares} highlight the issues of incomplete tooth models, misalignment within the same jaw, and the appearance of triangular patterns. Our method incorporates data serialization, where each shift window selects a subset of the point cloud that maintains the same relative positions across different teeth. This ensures accurate recognition of inter-tooth relative positions even when the tooth point cloud model is incomplete, resulting in better handling of wisdom teeth and tooth misalignment issues.

\subsubsection{Comparison with central point information}

\begin{figure}
    \centering
    \includegraphics[width=1.0\textwidth]{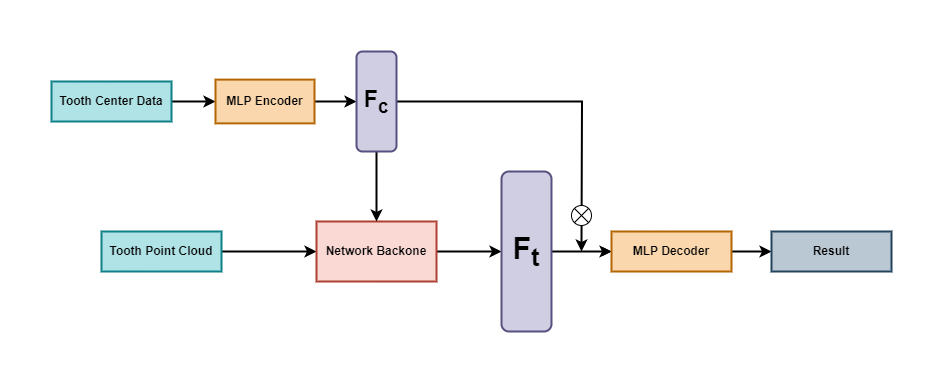}
    \caption{Network structure diagram with tooth center point feature encoding added when using other tooth alignment methods. After encoding the tooth center features, they are incorporated into the backbone network of other methods, and the residual of the tooth center features is added to the output of the backbone network.}
    \label{compare_net}
\end{figure}

The existing STAT methods do not directly use center point information as we do in this study. To ensure the validity of the experimental comparisons, the performance of each method was also tested with the addition of Tooth Center Data. For this purpose, we designed comparative networks that input both the tooth point cloud information (Tooth Point Cloud in Figure \ref{compare_net}) and the tooth center information (Tooth Center Data in Figure \ref{compare_net}). As shown in Figure \ref{compare_net}, the network uses the primary backbone model of these STAT methods to encode the point cloud into features \( F_t \). Then, an MLP-based tooth center point feature encoding module is added, and the center point latent vector \( F_c \) extracted by this module is added as a residual to the feature transmission of \( F_t \). Finally, a unified approach is used to regress the rotation and translation results.

The test results, as shown in Table \ref{w_wo_center}, indicate that except for a slight improvement in TAligNet, the addition of tooth center information did not significantly benefit the other network structures. In contrast, our network structure showed the highest improvement. This is because the SWTBS module's emphasis on center point features further enhances the focus on the overall arrangement, compensating for SWTP's neglect of center point information. Incorporating center point regression, our method achieved the most accurate results.

\begin{table}
	\caption{The impact of using tooth center feature regression on all alignment methods. ETL represents the number of training epochs when the total loss first converges below 10.0, indicating the convergence speed during training. The test AUC metric represents the accuracy of the final results.}
	\centering
	\begin{tabular}{lllll}
		\toprule
		\multirow{2}{*}{Model} & \multicolumn{2}{c}{without center info} & \multicolumn{2}{c}{with center info} \\
		\cline{2-5}
		 & ETL=10 $\downarrow$  & ADD/AUC $\uparrow$ & ETL=10 $\downarrow$ & ADD/AUC $\uparrow$ \\
		\midrule
		TAligNet       & 104 & 0.77 & 111 & 0.81 \\
		TANet          & 125 & 0.82 & 122 & 0.82 \\
		PSTN           & 112 & 0.79 & 105 & 0.78 \\
		Ptv3           & 121 & 0.82 & 134 & 0.82 \\
		Ours           & \textbf{129} & \textbf{0.74} & \textbf{122} & \textbf{0.90} \\
		\bottomrule
	\end{tabular}
	\label{w_wo_center}
\end{table}

\subsection{Visual results with complex cases}

\begin{figure}
    \centering
    \includegraphics[width=1.0\textwidth]{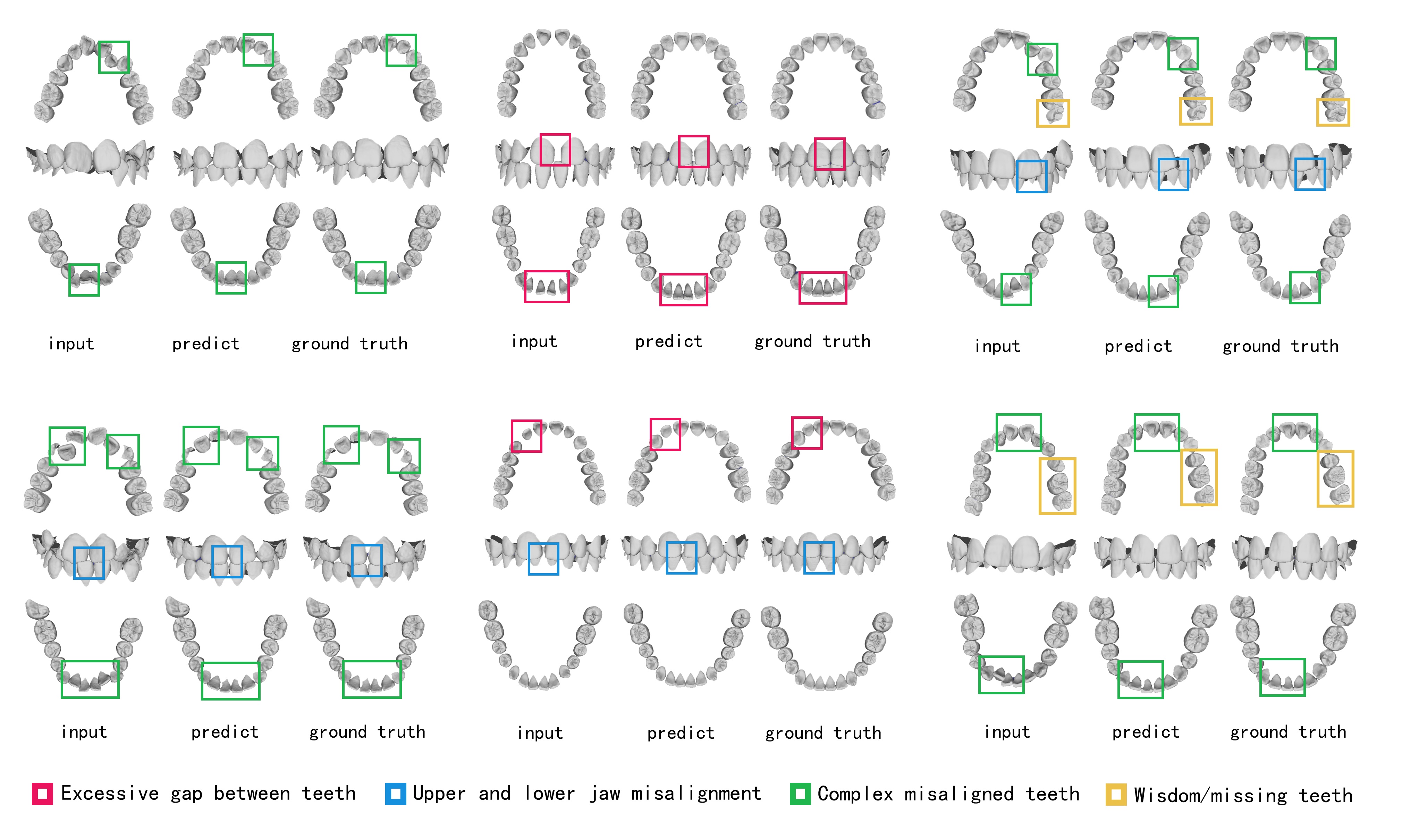}
    \caption{The figure shows the alignment prediction results of our method on 6 data cases, divided into 6 blocks. Each block has three columns: from left to right are the initial input data, prediction results, and ground truth. Each block has three rows, showing the corresponding data in the upper occlusal view, frontal view, and lower occlusal view. The colored boxes in the figure indicate different issues: red boxes represent large gaps, blue boxes represent occlusal misalignment, green boxes represent cross-bite, and yellow boxes represent wisdom teeth/missing teeth.}
    \label{effect_n}
\end{figure}

Figure \ref{effect_n} shows a comparison of our tooth alignment results. It can be seen that our method produces very neat alignments even in complex situations such as large gaps, crossbite, and triangular tooth arrangements. The introduction of serialization enhances the transformer's ability to perceive the positions of teeth within the jaw, and the shift window efficiently extracts local features of the teeth, supported by a comprehensive combination of loss functions. Additionally, our network specifies a maximum of 16 teeth per jaw, allowing it to accommodate cases where there are wisdom teeth on both sides or missing teeth.

\subsection{Ablation study}

To further demonstrate the effectiveness of the network architecture and the supervised learning performance of the loss functions, we conducted a series of ablation experiments. These experiments included analyses of the loss functions, network modules, point cloud serialization, data augmentation, and iterative training.

\subsubsection{Loss functions}

In the Section \ref{section:Methodology}, we discussed the occlusal distance uniformity loss function, the occlusal projection range alignment loss function, and the loss functions for rotation and translation transformation parameters. Below are the results validating the effectiveness of each loss through ablation experiments. The model reconstruction loss and transformation parameter loss, introduced in other methods, are highly effective in improving accuracy, as they emphasize the discrepancy between the predicted results and the ground truth from different perspectives. As shown in the first three rows of Table \ref{loss_func_abl}, our experiments also confirm that both types of losses need to be used simultaneously.

The occlusal projection range alignment loss and the occlusal distance uniformity loss proposed in this paper restrict the occlusion of the upper and lower teeth based on oral medical principles, ensuring that the upper and lower jaws are closely aligned according to natural occlusion laws and gradually move to their correct positions. Although the addition of these two losses increases the training time, it significantly improves prediction accuracy. The last three rows of Table \ref{loss_func_abl} present the ablation experiments for these two loss functions, confirming their effectiveness.

\begin{table}
	\caption{Ablation experiment results of loss functions, testing the impact of different loss function combinations on final test results.}
	\centering
	\begin{tabular}{llll}
		\toprule
		\multirow{2}{*}{Loss fuc} & \multicolumn{3}{c}{Test result} \\
		\cline{2-4}
		& $ADD/AUC \uparrow$ & $ME_{rotate} \downarrow$ & $ME_{translate} \downarrow$ \\
		\midrule
		$L_{recon}$                               & 0.64 & 9.6  & 2.7 \\
		$L_{val}$                                 & 0.62 & 10.5 & 3.1 \\
		$L_{recon} + L_{val}$                     & 0.79 & 8.3  & 2.2 \\
		$L_{recon} + L_{val} + L_{uni}$           & 0.81 & 5.9  & 1.7 \\
		$L_{recon} + L_{val} + L_{fit}$           & 0.83 & 5.3  & 1.4 \\
		$L_{recon} + L_{val} + L_{uni} + L_{fit}$ & \textbf{0.89} & \textbf{2.7}  & \textbf{1.1} \\
		\bottomrule
	\end{tabular}
	\label{loss_func_abl}
\end{table}

\subsubsection{Network Architecture}

In extracting inter-tooth features, we employed the multi-stage feature fusion architecture (SWTP) from the Swin-T paper. Additionally, we used the ST block-based tooth center feature extraction module (SWTBS) to further emphasize the extraction of relative inter-tooth position features. To validate the effectiveness of our network structure, we performed ablation experiments by dismantling or replacing various network modules.

While maintaining the functionality of SWTP, we tested replacements for SWTBS, including VTBS (using Vision Transformer blocks), SWTBS v2 (using Swin Transformer blocks v2), and a setup without the tooth center feature module. As shown in Table \ref{net_arch_abl}, the results demonstrate that using the tooth center feature module yields better performance compared to not using it, with SWTBS outperforming both VTBS and SWTBS v2. The Swin-T block's sliding window and merging mechanisms are more effective for feature fusion in tooth point clouds, whereas VTPS's Vision-T block lacks such mechanisms. Moreover, SWTBS v2's Swin-T v2 block's window movement method reduces prediction regression accuracy, a problem not encountered with the Swin-T block.

The latter part of Table \ref{net_arch_abl} records the results of the aforementioned structures when SWTP is removed. Without SWTP, the accuracy of the various tooth center feature modules remains relatively consistent but is lower compared to when SWTP is included. This is because SWTP's multi-stage network architecture efficiently extracts global features between independent point clouds, offering a more rational field of view expansion process compared to traditional visual transformers. Additionally, we also examined the performance of using only PTv1, which shows greater rotational error compared to using only SWTP. This is due to PTv1 lacking tools like sliding windows that enhance relative inter-tooth position features, making its generic point cloud feature extraction less effective than the multi-stage structure specifically designed for jaw point clouds.

\begin{table}
	\caption{Ablation experiment results of network architecture. The prediction accuracy of different network structures was tested, including the block composition of other transformer methods and the use of the SWTP module.}
	\centering
	\begin{tabular}{lllll}
		\toprule
		Methods & w/o SWTP & $ADD/AUC \uparrow$ & $ME_{rotate} \downarrow$ & $ME_{translate} \downarrow$ \\
		\midrule
		VTBS         & \usym{1F5F8} & 0.79  & 7.10 & 1.83 \\
		SWTBS v2     & \usym{1F5F8} & 0.85  & 5.60 & 1.40 \\
		PTv3         & \usym{1F5F8} & /     & /   & /     \\
		SWTBS(Ours)  & \usym{1F5F8} & \textbf{0.90}  & \textbf{2.70} & \textbf{1.10} \\
		\midrule
		VTBS         & \usym{2715}  & 0.75  & 8.50 & 2.20 \\
		SWTBS v2     & \usym{2715}  & 0.78  & 7.80 & 2.10 \\
		PTv3         & \usym{2715}  & 0.73  & 9.80 & 2.40 \\
		SWTBS(Ours)  & \usym{2715}  & 0.81  & 7.20 & 2.00 \\
		\bottomrule
	\end{tabular}
	\label{net_arch_abl}
\end{table}

It can be observed that compared to network structures using only Swin blocks or Vision blocks, the multi-stage Swin block structure yields higher accuracy. This is because the multi-stage approach reduces the size of the latent vector progressively through dimensional merging, which is more effective in retaining task-specific dental features than directly passing down through averaging.

\subsubsection{Point Cloud Serialization}

In Section \ref{section:Methodology}, we discussed sorting the point cloud data of individual teeth for input into the Swin-T multi-layer feature fusion module. Serialization, if it ensures that the point cloud data selected by the window corresponds to the same local region of the teeth, can better extract the relative position features between teeth \cite{wu2023point}. To this end, we proposed a serialization method based on the simulated dental arch curve. To verify its effectiveness, we compared it with three other sorting methods (two ordered and one random). The statistical data are presented in Table \ref{pc_ser_abl}.

\begin{table}
	\caption{Ablation experiment results of point cloud serialization methods, exploring the impact of four different point cloud ordering strategies on the final results.}
	\centering
	\begin{tabular}{llll}
		\toprule
		\multirow{2}{*}{Serialization Function} & \multicolumn{3}{c}{Test result} \\
		\cline{2-4}
		& $ADD/AUC \uparrow$ & $ME_{rotate} \downarrow$ & $ME_{translate} \downarrow$ \\
		\midrule
		Random Order                 & 0.77 & 6.1 & 1.9 \\
		Based on dental local z-axis & 0.80 & 5.4 & 1.7 \\
		Based on dental arch center  & 0.82 & 5.6 & 1.3 \\
		Based on virtual arch line   & \textbf{0.89} & \textbf{2.7} & \textbf{1.1} \\
		\bottomrule
	\end{tabular}
	\label{pc_ser_abl}
\end{table}

As shown in the Table \ref{pc_ser_abl}, the random sorting method yielded the worst results, as it completely failed to leverage the sequential information to enhance the transformer's performance. Sorting based on the local z-axis of the teeth provided some sequence-related benefits, resulting in relatively better performance, but it was still insufficient due to the common occurrence of multi-peaked tooth crowns, particularly in the posterior teeth, which may cause sequentially arranged points to be located in different local regions. The center-point-based sorting method addressed this issue but still introduced unavoidable angular deviations for posterior teeth, as the arrangement of teeth is typically U-shaped rather than O-shaped. Our proposed method, which is based on the simulated dental arch curve for U-shaped jaw structures, achieved better prediction results, as shown in Table \ref{pc_ser_abl}.

\subsubsection{Data Augmentation}

As described in Section \ref{subsection:Constraint Data augmentation}, to mitigate the shortcomings of insufficient data, we employed a data augmentation method constrained by medical rules. However, more data augmentation is not always better; a high proportion of augmented data can lead to network distortion. To explore the optimal level of data augmentation, we conducted ablation experiments using two methods: regular augmentation (during the training phase) and constrained augmentation (during the preprocessing phase). These two augmentation methods were not used simultaneously; instead, one of them was randomly selected based on a probability \( \rho \).

\begin{figure}
    \centering
    \subfigure{
        \includegraphics[width=0.45\textwidth]{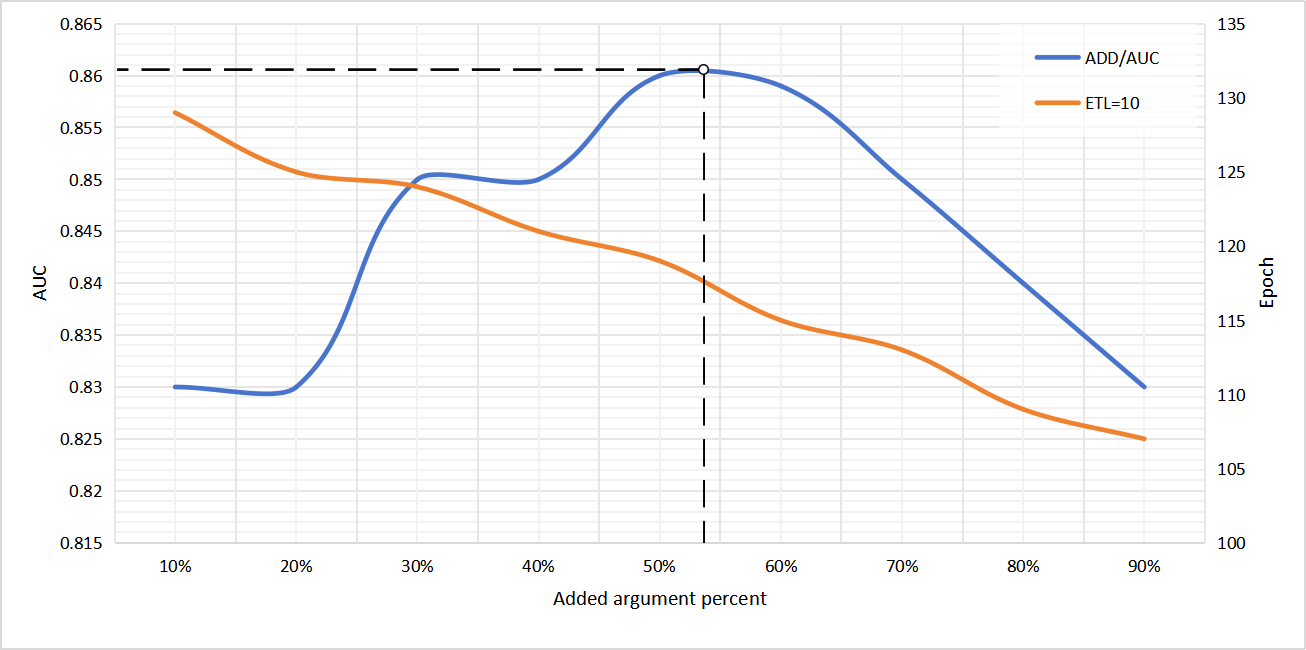}
        \label{aug_strong}
    }
    \subfigure{
        \includegraphics[width=0.45\textwidth]{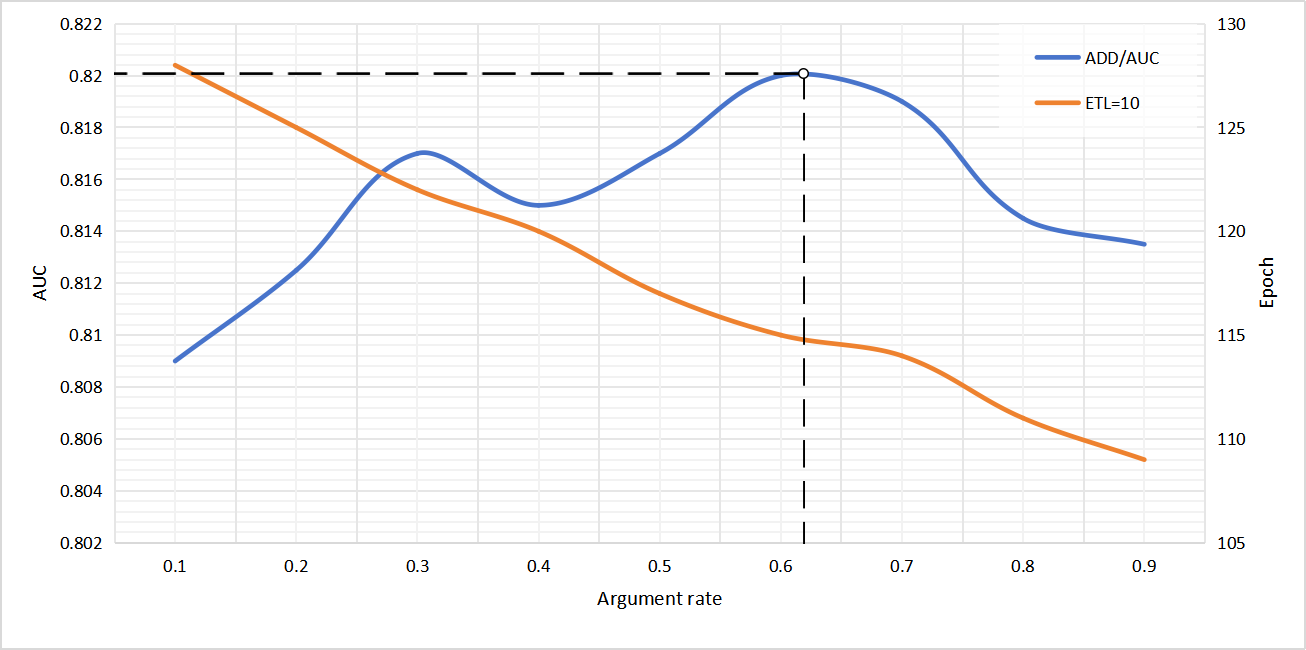}
        \label{aug_fast}
    }
    \caption{The effect of data augmentation intensity on final prediction accuracy and training convergence speed. The left side shows constrained augmentation, where the independent variable is the ratio of augmented data to the original total data quantity. The right side shows standard augmentation, where the independent variable is the probability of augmentation triggering during training.}
    \label{aug_chart}
\end{figure}

To estimate the optimal parameter \( \rho \), we adjusted the proportion of data augmented by constraint augmentation within the dataset and conducted ablation experiments with various proportions of augmented data. We also performed ablation experiments by varying the probability of ordinary augmentation. According to experimental protocol, each experiment used only one augmentation method for training and testing. Figure \ref{aug_chart} illustrates the relationship between the degree of augmentation and training convergence speed, as well as prediction accuracy. Here, ETL represents the total training loss converging to within 10.0, and the number of epochs trained. Although convergence speed improves with augmentation, the proportion of augmented data does not linearly correlate with the final results. For our network, appropriate data augmentation can enhance prediction accuracy; however, excessive augmentation can degrade the results, as simulated data does not fully replicate real data. Figure \ref{aug_chart} shows that the peak prediction accuracy is achieved with a constraint augmentation ratio of 54$\%$ and a ordinary augmentation trigger probability of 62$\%$.

The two data augmentation methods are relatively independent but are applied during different phases: constraint augmentation is implemented during preprocessing, while ordinary augmentation is applied during training. Therefore, these methods can be used together, and we conducted experiments with both augmentation methods applied simultaneously, as shown in table \ref{aug_use_abl}. In the comparative experiments, both constraint augmentation and ordinary augmentation were used at the levels providing the highest accuracy, with constraint augmentation data comprising 54$\%$ of the original training dataset and a ordinary augmentation trigger probability set at 62$\%$.

\begin{table}
	\caption{Ablation experiment results of data augmentation, recording the impact of using pre-orthodontic and post-orthodontic data for augmentation, as well as the effect of including constraints on the final experimental results.}
	\centering
	\begin{tabular}{lllll}
		\toprule
		\multicolumn{2}{c}{Augmentation Way} & $ADD/AUC \uparrow$ & $ETL=10 \downarrow$ & $SigAugTime \downarrow$ \\
		\midrule
		\multicolumn{2}{c}{None Augmentation}                & 0.83 & 145 & /    \\
		\multirow{2}{*}{Source Data} & With Constraint    & 0.86 & 141 & 3.53 \\
		                                     & Without Constraint & 0.84 & 125 & 0.84 \\
		\multirow{2}{*}{Target Data} & With Constraint    & \textbf{0.90} & \textbf{134} & \textbf{3.51} \\
		                                     & Without Constraint & 0.85 & 109 & 0.84 \\
		\bottomrule
	\end{tabular}
	\label{aug_use_abl}
\end{table}

constraint augmentation during the data preprocessing phase contributes more significantly to improving experimental results. Ordinary augmentation, due to its higher level of data simulation, deviates more from real data, resulting in relatively lesser improvement. The experimental results combining both methods are essentially consistent with those obtained using only constraint augmentation, as both methods affect testing accuracy through similar principles by enriching data samples to expand the learning scope. Therefore, when preprocessing time is sufficient, constraint augmentation can be used independently. Conversely, if prediction accuracy requirements are less stringent and faster runtime is needed, ordinary data augmentation can be used on its own.

\subsubsection{Iteration Experiment}

Prediction errors are inherent in network models. In this paper, we explored the use of iterative testing to improve prediction accuracy\cite{tanet2020}\cite{lei2023automatic}. Specifically, the network's prediction results for the test data are treated as new initial data, which is then fed back into the network for iterative prediction. We experimented with the effects of multiple iterations using our method, as shown in Table \ref{iter_exp_abl}. It can be observed that the accuracy of results improves during the first two iterations. However, as the number of iterations increases, the error stabilizes. This is because, after a certain number of iterations, the network structure reaches its accuracy limit, and further iterations do not yield more accurate results, thus causing the error to remain constant.

\begin{table}
	\caption{Iterative experiment tests, where the experimental accuracy remains unchanged after multiple iterations, indicating that the network's learning of the data has reached its limit.}
	\centering
	\begin{tabular}{llll}
		\toprule
		\multirow{2}{*}{Iterations} & \multicolumn{3}{c}{Test result} \\
		\cline{2-4}
		& $ADD/AUC \uparrow$ & $ME_{rotate} \downarrow$ & $ME_{translate} \downarrow$ \\
		\midrule
		1 & 0.889 & 2.804 & 1.156 \\
		2 & 0.892 & 2.784 & 1.179 \\
		3 & 0.903 & 2.761 & 1.045 \\
		4 & 0.901 & 2.786 & 1.108 \\
		5 & 0.899 & 2.792 & 1.154 \\
		6 & 0.903 & 2.735 & 1.126 \\
		\bottomrule
	\end{tabular}
	\label{iter_exp_abl}
\end{table}

\section*{Conclusions and Future Work}

\subsection*{Conclusion}
This paper proposes a novel, high-precision, and efficient neural network approach for tooth alignment prediction. It is the first in the field of tooth arrangement to use the multi-level feature fusion structure of Swin-T as its core, supplemented by a tooth center feature extraction module that emphasizes global features. This method improves upon and draws inspiration from existing data augmentation mechanisms, proposing a constrained augmentation method that enhances the authenticity of the augmented data. Additionally, the paper designs two occlusion evaluation loss functions that effectively describe the specific occlusion relationships between upper and lower teeth, achieving remarkable results in the task of tooth arrangement in orthodontic work.

Through comparative experiments and ablation studies, the effectiveness of each module of this method was verified, achieving extremely high prediction accuracy in the field of tooth arrangement. The study also explored the impact of data augmentation ratios and multiple iterations on the proposed method. Furthermore, this paper constructed the first open dataset in the field of tooth arrangement: \textbf{OrthData2024}. This dataset includes over six hundred fully annotated data pairs, consisting of point clouds sampled from tooth crowns, addressing the issue of a lack of public testing datasets in this field.

\subsection*{Future work}
At present, dental alignment work only focuses on predicting the ideal outcome of orthodontic treatment. In fact, the orthodontic prediction that dentists currently need is to predict each step of the treatment process, not just the final effect. The iterative nature of the work in this article can be utilized to further modify and optimize the network, limiting the range and degree of rotation and translation changes in each iteration, and obtaining a process that gradually achieves ideal alignment of the gingival arch. Let each iteration represent a stage of orthodontic effect and gradually predict the final ideal orthodontic effect, which can better assist doctors in designing the entire treatment plan.

If there are teeth with severe lateral distortion, resulting in the teeth being closer to the centerline of the dental arch, it will have a certain impact on serialization and lead to inaccurate serialization. In the future, the serialization method can be improved to ensure that the sorting of the model is not affected by the position of the model in the world coordinate system.

\bibliographystyle{elsarticle-num}
\bibliography{./source/ref}

\end{document}